\documentclass[10pt,twocolumn,letterpaper]{article}

\usepackage{wacv}              % To produce the CAMERA-READY version
\usepackage{graphicx}
\usepackage{amsmath}
\usepackage{amssymb}
\usepackage{booktabs}
\usepackage{multirow}
\usepackage{enumitem}

\usepackage[pagebackref,breaklinks,colorlinks]{hyperref}

\usepackage[capitalize]{cleveref}
\crefname{section}{Sec.}{Secs.}
\Crefname{section}{Section}{Sections}
\Crefname{table}{Table}{Tables}
\crefname{table}{Tab.}{Tabs.}

\def\wacvPaperID{1229} % *** Enter the WACV Paper ID here
\def\confName{WACV}
\def\confYear{2025}

\begin{document}

\title{Guardian of the Ensembles: Introducing \underline{P}airwise \underline{A}dversarially \underline{R}obust \underline{L}oss for Resisting Adversarial Attacks in DNN Ensembles}

\author{
Shubhi Shukla\textsuperscript{1}, Subhadeep Dalui\textsuperscript{2}, Manaar Alam\textsuperscript{3}, Shubhajit Datta\textsuperscript{4},\\
Arijit Mondal\textsuperscript{5}, Debdeep Mukhopadhyay\textsuperscript{2}, Partha Pratim Chakrabarti\textsuperscript{2}\\
\textsuperscript{1}Centre for Computational and Data Sciences, IIT Kharagpur, India\\
\textsuperscript{2}Computer Science and Engineering Department, IIT Kharagpur, India\\
\textsuperscript{3}Center for Cyber Security, New York University Abu Dhabi, UAE\\
\textsuperscript{4}Department of Artificial Intelligence, IIT Kharagpur, India\\
\textsuperscript{5}Computer Science and Engineering Department, IIT Patna, India\\
\tt\small \{shubhishukla, csesubhadeep2022\}@kgpian.iitkgp.ac.in, alam.manaar@nyu.edu,  \\ 
\tt\small shubhajitdatta1988@gmail.com, arijit@iitp.ac.in, \{debdeep, ppchak\}@cse.iitkgp.ac.in\\
\textbf{Accepted at WACV 2025}
}
\maketitle

%%%%%%%%% ABSTRACT
\begin{abstract}
  Adversarial attacks rely on \textit{transferability}, where an \textit{adversarial example} (AE) crafted on a \textit{surrogate classifier} tends to mislead a \textit{target classifier}. Recent ensemble methods demonstrate that AEs are less likely to mislead multiple classifiers in an ensemble. This paper proposes a new ensemble training using a \texttt{P}airwise \texttt{A}dversarially \texttt{R}obust \texttt{L}oss (PARL) that \textit{by construction} produces an ensemble of classifiers with \textit{diverse decision boundaries}. PARL~utilizes outputs and gradients of each layer with respect to network parameters in every classifier within the ensemble simultaneously. PARL~is demonstrated to achieve higher robustness against \textit{black-box transfer attacks} than previous ensemble methods as well as adversarial training without adversely affecting clean example accuracy. Extensive experiments using standard Resnet20, WideResnet28-10 classifiers demonstrate the robustness of PARL~against state-of-the-art adversarial attacks. While maintaining similar clean accuracy and lesser training time, the proposed architecture has a $24.8\%$ increase in robust accuracy ($\epsilon = 0.07$) from the state-of-the art method. Code is available at: \url{https://github.com/shubhishukla10/PARL}
\end{abstract}

%%%%%%%%% BODY TEXT
\section{Introduction}

\label{sec:intro}

While Deep Learning (DL) models are extremely efficient in solving complicated decision-making tasks, they are vulnerable to well-crafted Adversarial Examples (AE)~\cite{DBLP:journals/corr/SzegedyZSBEGF13}. The widely-studied phenomenon of AE has produced numerous attacks with varied complexities and effective deceiving strategies~\cite{DBLP:journals/caaitrit/ChakrabortyADCM21}. An extensive spectrum of defenses against such attacks has also been proposed in the literature~\cite{DBLP:journals/caaitrit/ChakrabortyADCM21}, which generally falls into two categories. The first category enhances the training strategy of DL models to make them less vulnerable to AE~\cite{DBLP:journals/corr/GuR14,DBLP:conf/sp/PapernotM0JS16}. However, it has been demonstrated that these defenses are not generalized for all varieties of AE but are constrained to specific categories~\cite{DBLP:conf/sp/Carlini017,DBLP:conf/icml/AthalyeC018}. The second category intends to detect AE by simply flagging them~\cite{DBLP:conf/iclr/MetzenGFB17,DBLP:conf/iccv/LiL17}. However, it has been illustrated with several experiments that these detection techniques could be efficiently bypassed by a strong adversary having partial or complete knowledge of the internal working procedure~\cite{DBLP:conf/ccs/Carlini017}.

While the approaches mentioned above deal with standalone models, in this paper, \textit{we utilize an ensemble of models to resist adversarial examples (AE)}. The notion of using \textit{diverse ensembles} to increase robustness against AE has recently gained popularity~\cite{NEURIPS2020_b86e8d03}. The primary motivation for using an ensemble-based defense with diverse decision boundaries is that if multiple models with similar \textit{decision boundaries} perform the same task, the \textit{transferability property} of deep learning models makes it easier for an adversary to misclassify all the models simultaneously using AE crafted on any of the models. However, it will be difficult for an adversary to misclassify multiple models simultaneously if they have diverse decision boundaries.

\noindent \textbf{Related Works: Ensemble-based Adversarial Defense:} 

\cite{DBLP:journals/corr/abs-1709-03423}~introduced ensemble-based defense against AE using various ad-hoc techniques such as different random initializations, different neural network structures, bagging the input data, and adding Gaussian noise while training. \cite{DBLP:conf/iclr/TramerKPGBM18}~proposed \textit{Ensemble Adversarial Training} that incorporates perturbed inputs transferred from other pre-trained models during \textit{adversarial training} to decouple AE generation from the parameters of primary model. However, these methods do not explicitly focus on incorporating diversity in the decision boundaries of the models within an ensemble. \cite{DBLP:journals/corr/abs-1901-09981}~proposed \textit{Diversity Training} of an ensemble of models with uncorrelated loss functions using \textit{Gradient Alignment Loss} (GAL) to reduce the dimension of adversarial sub-space shared between different models and increase the robustness of the classification task. \cite{DBLP:conf/icml/PangXDCZ19}~proposed \textit{Adaptive Diversity Promoting} (ADP) regularizer to train an ensemble of models that encourages the non-maximal predictions in each member in the ensemble to be mutually orthogonal, degenerating the transferability that aids in resisting AE. \cite{DBLP:conf/nips/YangZDIGTWB020}~proposed a methodology, called DVERGE, that isolates the adversarial vulnerability in each model of an ensemble by distilling non-robust input features. \cite{DBLP:conf/nips/YangLXZCZRZL21}~proposed \textit{Transferability Reduced Smooth} (TRS) ensemble that enforces diversity among the models within an ensemble by simultaneously reducing loss gradient and smoothing decision regions using support instances as regularizers. Recently, \cite{DBLP:conf/aaai/0003NY023}~proposed \textit{Ensemble-in-One} (EIO), method which works by using a random gated network to exponentially increase the number of paths for ensemble learning within a single model, resulting in better adversarial robustness.

The methods mentioned above either do not inherently enforce diversity on decision boundaries of the models or are less robust against stronger adversaries, significantly impacting clean example accuracy. \textit{In this work, we propose a systematic approach that incorporates diversity among decision boundaries of submodels in an ensemble to enhance robustness against adversarial examples (AE). This diversity is achieved by considering mutual dissimilarity in gradients of each layer with respect to intermediate network parameters and the output of intermediate convolution layers during training. Such diversity reduces the transferability of AE within the ensemble.}

\noindent \textbf{Intuition behind the Proposed Approach: }

\begin{figure}[!t]
\centering
\subfloat[\label{fig:motivation:input_image}]{\includegraphics[width=0.2\linewidth]{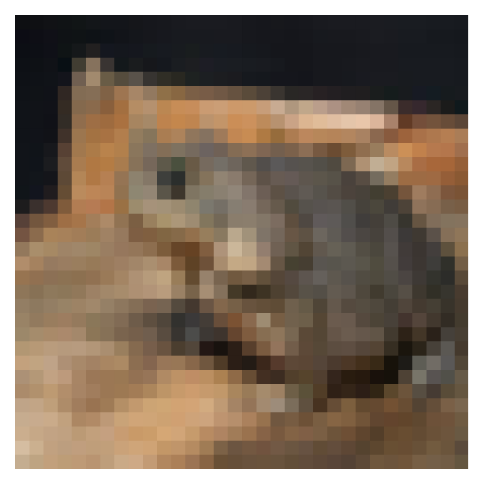}}
\subfloat[\label{fig:motivation:primary_gradient}]{\includegraphics[width=0.2\linewidth]{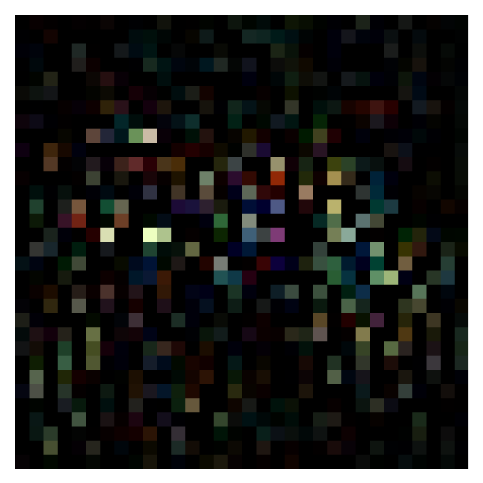}}
\subfloat[\label{fig:motivation:similar_gradient}]{\includegraphics[width=0.2\linewidth]{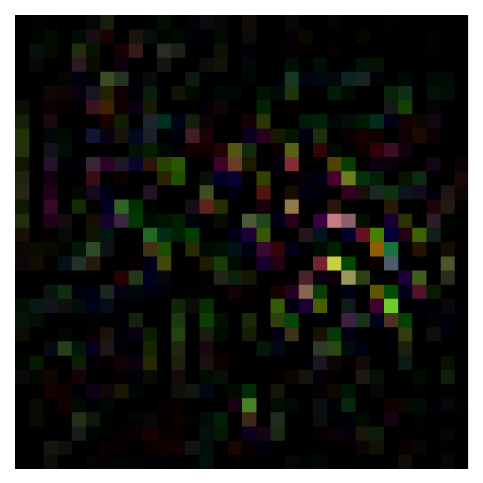}}
\subfloat[\label{fig:motivation:diverse_gradient}]{\includegraphics[width=0.2\linewidth]{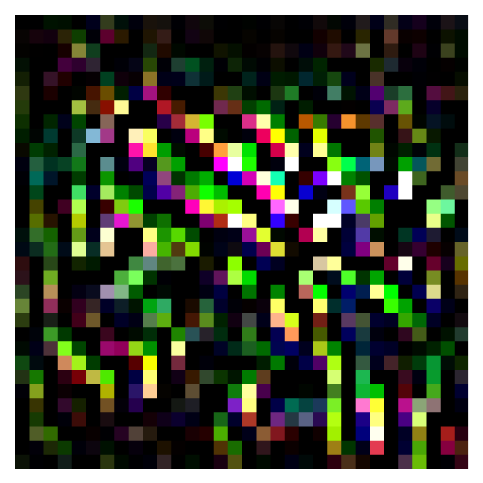}}
\subfloat[\label{fig:motivation:angle}]{\includegraphics[width=0.2\linewidth, height=1.5cm]{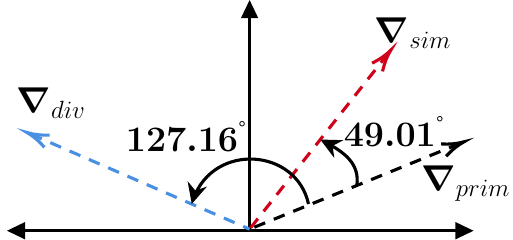}}
\vspace{-0.3cm}
\caption{\footnotesize (a) Input image; (b) $\nabla_{prim}$: Gradient of loss in the primary model; (c) $\nabla_{sim}$: Gradient of loss in another model with \textit{similar} decision boundaries; (d) $\nabla_{div}$: Gradient of loss in a model with \textit{not so similar} decision boundaries but comparable accuracy; (e) Symbolic directions of all the gradients in higher dimensions. Gradients are computed with respect to the image shown in (a). \vspace{-0.6cm}}
\label{fig:motivation}
\end{figure}

The first part our proposed ensemble loss method aims to diversify classifiers by making their gradients dissimilar/orthogonal. We illustrate this with an example of image classifiers trained on CIFAR-10, shown in Fig.~\ref{fig:motivation}.

Fig.~\ref{fig:motivation:input_image} shows an input image of a `frog' which we use to demonstrate how gradient of loss with respect to intermediate convolution layer parameters is visualized in classifiers $\mathcal{M}_{prim}$ and $\mathcal{M}_{sim}$, which have similar decision boundaries, and in $\mathcal{M}_{div}$, which has a distinctly different decision boundary from $\mathcal{M}_{prim}$. Although $\mathcal{M}_{prim}$ and $\mathcal{M}_{sim}$ are trained under the same settings but with different initializations, their gradients, $\nabla_{prim}$ (Fig.~\ref{fig:motivation:primary_gradient}) and $\nabla_{sim}$ (Fig.~\ref{fig:motivation:similar_gradient}), tend to point in nearly the same directions as shown in Fig.~\ref{fig:motivation:angle}. In contrast, the gradient of $\mathcal{M}_{div}$, $\nabla_{div}$ (Fig.~\ref{fig:motivation:diverse_gradient}), significantly diverges. This indicates that adversarial examples crafted for $\mathcal{M}_{prim}$ can easily fool $\mathcal{M}_{sim}$ but are less likely to affect $\mathcal{M}_{div}$, highlighting the impact of decision boundary diversity on adversarial example transferability.
Focusing on gradients relative to intermediate layer weights rather than just the input delves deeper into the neural network's learning process. This helps in targeting the core of the classifiers' feature extraction and representation mechanisms. Early and intermediate convolutional layers are where raw input data begins its transformation into a hierarchy of features, which are then used for classification. By promoting orthogonal gradients in these layers, we ensure that each classifier within the ensemble develops a unique approach to processing and interpreting the input data, leading to diverse feature representations.

\begin{figure}[!t]
\centering
\subfloat[\label{fig:motivation_corr:input_image}]{\includegraphics[width=0.2\linewidth]{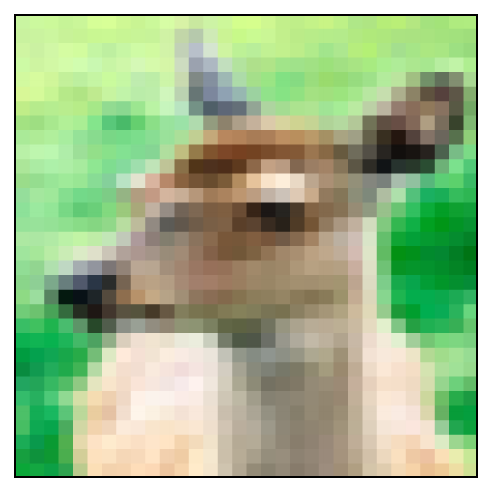}}
\subfloat[\label{fig:motivation_corr:surrogate1}]{\includegraphics[width=0.2\linewidth]{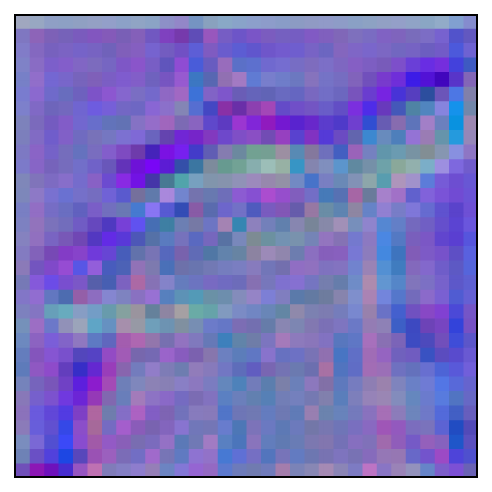}}
\subfloat[\label{fig:motivation_corr:surrogate2}]{\includegraphics[width=0.2\linewidth]{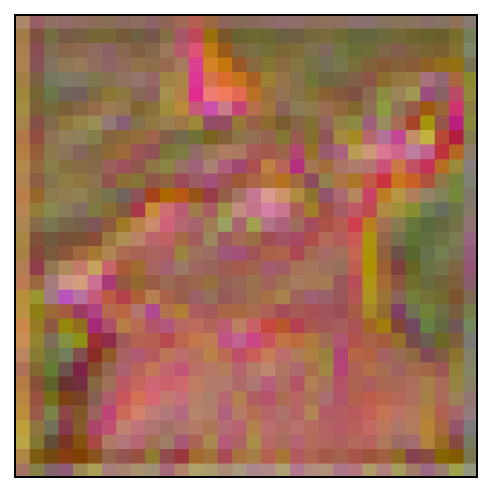}}
\subfloat[\label{fig:motivation_corr:parl1}]{\includegraphics[width=0.2\linewidth]{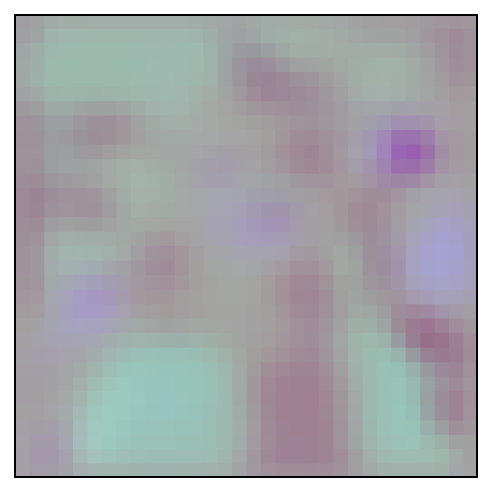}}
\subfloat[\label{fig:motivation_corr:parl2}]{\includegraphics[width=0.2\linewidth]{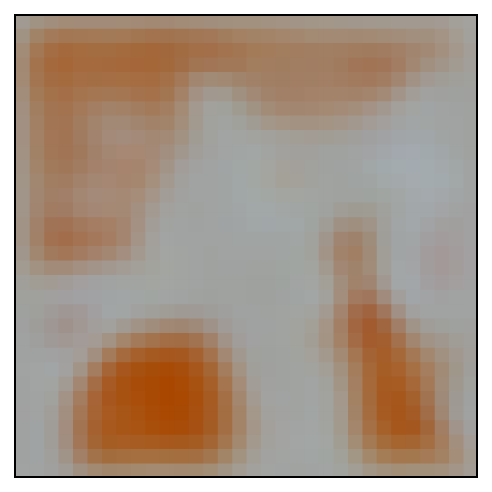}}
\vspace{-0.3cm}
\caption{\footnotesize (a) Input image; (b) and (c) Similar outputs of an intermediate convolution layer of two classifiers with similar internal representations; (d) and (e) Contrasting outputs of an intermediate convolution layer of two classifiers which are trained simultaneously to have distinct internal representations \vspace{-0.6cm}}
\label{fig:motivation_corr}
\end{figure}

Building on the gradient diversity objective, our ensemble loss method aims to further diversify internal representations by minimizing the correlation between outputs of intermediate convolutional layers across classifiers. Unlike cosine similarity, which encourages gradient orthogonality, correlation is used here to assess similarity between intermediate layer outputs. This is because convolutional layers capture spatial hierarchies of features where both activation patterns (direction) and intensities (magnitude) are crucial. Correlation accounts for both aspects, providing a comprehensive measure of similarity and ensuring each classifier uniquely contributes to the ensemble, thus increasing robustness against adversarial attacks. Fig.\ref{fig:motivation_corr} illustrates this with an input image labeled as ‘Deer’ from CIFAR-10, processed by four CNN classifiers. The first two classifiers, trained identically but with different initializations, produce similar outputs (Fig.\ref{fig:motivation_corr:surrogate1} and Fig.\ref{fig:motivation_corr:surrogate2}). The latter two classifiers, specifically trained for distinct representations, show significantly different outputs (Fig.\ref{fig:motivation_corr:parl1} and Fig.~\ref{fig:motivation_corr:parl2}), demonstrating enforced diversity.

\textit{Our method is among the first to encourage diversity at both the decision boundary and intermediate representation levels. This novel approach ensures a more robust model compared to previous adversarial ensemble defenses by diversifying the potential paths an adversarial input might take, making it harder for such attacks to succeed.}

% }

\noindent \textbf{Contributions:} We summarize our contributions below:
\begin{itemize}[itemsep=0.1pt]
    \item We propose a method that, \textit{by construction}, increases diversity in the decision boundaries among all the models within an ensemble to degrade the transferability of AE.
    \item We propose a \textit{Pairwise Adversarially Robust Loss} (PARL) function by utilizing outputs and gradients of each layer of every model within the ensemble simultaneously while training to produce such varying decision boundaries.
    \item PARL significantly improves the overall robustness of an ensemble against black-box transfer attacks without substantially impacting the clean example accuracy.
\end{itemize}

We evaluated PARL extensively using CIFAR-10, CIFAR-100, and Tiny Imagenet datasets with \textit{Resnet20} and \textit{WideResnet28-10} architectures against state-of-the-art adversarial attacks such as \textit{PGD}\cite{DBLP:conf/iclr/MadryMSTV18}, \textit{M-DI$^2$-FGSM}\cite{DBLP:conf/cvpr/XieZZBWRY19}, \textit{SGM}\cite{DBLP:conf/iclr/Wu0X0M20}, and \textit{Square}\cite{DBLP:conf/eccv/AndriushchenkoC20}. We compared PARL with previous ensemble adversarial defenses and the recent adversarial training method \textit{TRADES}. At the highest perturbation strength of 0.07, PARL achieved a robust accuracy surpassing the state-of-the-art ensemble defense by 24.8\%, with nearly one-third of the training time and similar clean accuracy for the CIFAR-10 Resnet20 model. Additionally, compared to TRADES, PARL showed similar robust accuracy and 3.68\% higher clean accuracy, offering a better balance of security and utility.

\section{Building Ensemble Networks using PARL}
\noindent \textbf{Threat Model:}
We define our threat model for generating AE in the context of a \emph{Zero Knowledge Adversary}. This adversary lacks access to the target ensemble $\mathcal{M}_{T}$, but possesses knowledge of a surrogate ensemble $\mathcal{M}_{S}$ that has been trained using the same dataset. Also known as a \textit{black-box} adversary, the adversary formulates AE on the source model $\mathcal{M}_{S}$ and subsequently transfers them to the target model $\mathcal{M}_{T}$.

\noindent \textbf{Overview of PARL:}
Consider an ensemble $\mathcal{M}_\mathcal{T}$ consisting of $\mathcal{N}$ neural networks and denoted as $\mathcal{M}_{\mathcal{T}} = \bigcup_{i=1}^{\mathcal{N}} \mathcal{M}_{i}$, where $\mathcal{M}_{i}$ is the $i^{th}$ network in the ensemble. All $\mathcal{M}_{i}$'s are trained simultaneously using PARL, which we subsequently discuss in detail. The final decision for an input on $\mathcal{M}_{\mathcal{T}}$ is decided over \textit{majority voting} among all $\mathcal{M}_{i}$'s. Formally, assume a test set of $t$ inputs $\{\mathbf{x}_1, \mathbf{x}_2, \dots, \mathbf{x}_t\}$ with respective ground truth labels as $\{\mathbf{y}_1, \mathbf{y}_2, \dots, \mathbf{y}_t\}$. The final decision of $\mathcal{M}_{\mathcal{T}}$ for an input $\mathbf{x}_j$ is defined as

\vspace{-\baselineskip} 
{\small
\begin{equation*}
\mathcal{C}(\mathcal{M}_{\mathcal{T}}, \mathbf{x}_j) = majority\{\mathcal{M}_{1}(\mathbf{x}_j), \mathcal{M}_{2}(\mathbf{x}_j), \cdots, \mathcal{M}_{\mathcal{N}}(\mathbf{x}_j)\}
\end{equation*}}

\noindent $\mathcal{C}(\mathcal{M}_{\mathcal{T}}, \mathbf{x}_j) = \mathbf{y}_j$ for most $\mathbf{x}_j$'s in an appropriately trained $\mathcal{M}_{\mathcal{T}}$. The primary argument behind PARL is that all $\mathcal{M}_{i}$'s have dissimilar decision boundaries but not significantly different accuracies. Hence, a clean example classified as class $\mathcal{C}_\mathbf{x}$ in $\mathcal{M}_i$ will also be classified as $\mathcal{C}_\mathbf{x}$ in most other $\mathcal{M}_{j}$'s (where $j=1 \dots \mathcal{N}$, $j \neq i$) with a high probability. Consequently, due to the diversity in decision boundaries between $\mathcal{M}_{i}$ and $\mathcal{M}_{j}$ (for $i, j =1 \dots \mathcal{N}$ and $i \neq j$), the AE generated for a surrogate ensemble $\mathcal{M}_\mathcal{S}$ will have a different impact on each classifiers within $\mathcal{M}_{\mathcal{T}}$, i.e., the \textit{transferability} of AE will be challenging within the ensemble. The adversary can also generate AE for $\mathcal{M}_{\mathcal{T}}$. However, the input image perturbation will be in different directions due to the diversity in decision boundaries among all $\mathcal{M}_{i}$'s. The collective disparity in perturbation directions makes it challenging to craft AE for the ensemble.

\noindent \textbf{Basic Terminologies used in PARL: }
We assume that each $\mathcal{M}_{i}$ in $\mathcal{M}_\mathcal{T}$ has the same architecture with $\mathcal{H}$ hidden layers. Let $\mathcal{J}_{\mathcal{M}_{i}}(\mathbf{x}, \mathbf{y})$ be the loss function for the network $\mathcal{M}_{i}$ considering a data point $\mathbf{x}$, where $\mathbf{y}$ is the ground-truth label for $\mathbf{x}$. Let $\mathcal{F}^{k}_{i}(\mathbf{x})$ be the output of $k^{th}$ hidden layer of $\mathcal{M}_{i}$ for input $\mathbf{x}$ and let ${w}^{k}_{i}$ denote all network parameters up to $k^{th}$ hidden layer which are involved in computation of $\mathcal{F}^{k}_{i}(\mathbf{x})$. \color{black}Let us consider $\mathcal{F}^{k}_{i}(\mathbf{x})$ has $\mathcal{D}_{k}$ number of output features. Let $\nabla_{{w}^{k}_{i}}\mathcal{F}^{k}_{i}(\mathbf{x})$ denote the sum of gradients over each output feature of $k^{th}$ hidden layer with respect to the parameters represented by ${w}^{k}_{i}$ on the network $\mathcal{M}_{i}$ for data point $\mathbf{x}$. Hence, $\nabla_{{w}^{k}{i}}\mathcal{F}^{k}{i}(\mathbf{x}) = \sum_{f=1}^{\mathcal{D}{k}}\nabla{{w}^{k}{i}}[\mathcal{F}^{k}{i}(\mathbf{x})]_{f}$

\noindent where $\nabla_{{w}^{k}_{i}}[\mathcal{F}^{k}_{i}(\cdot)]_{f}$ is gradient of $f^{th}$ output feature of $k^{th}$ hidden layer on network $\mathcal{M}_{i}$ with respect to the parameters represented by ${w}^{k}_{i}$ for data point $\mathbf{x}$. Let $\mathcal{X}$ be the training dataset with $|\mathcal{X}|$ examples.

\noindent \textbf{PARL Construction: }
The main idea behind PARL is to train an ensemble of neural networks with diverse decision boundaries. To achieve such diversity the parameters in each network which are learned during training must be dissimilar across the ensemble. In this paper, we introduce PARL to train an ensemble so that the gradients of loss with respect to the network parameters lead to different directions in different networks for the same input. The gradients guide training of any neural network by giving an idea of the direction in which the parameters should be updated. Hence, the fundamental strategy is to make these gradients as dissimilar as possible while training all the networks simultaneously. The loss is computed using the output of the last layer and ground truth label. However, the last layer output depends on all intermediate layers' outputs. Therefore, loss and so its gradient both depend on the intermediate layers' outputs. As a result, employing diversity in intermediate layers will also enforce diversity in the model decision boundary. Hence, instead of loss, we considered the output of intermediate layers for implementing diversity with a higher degree of control and better flexibility in employing constraints. Recognizing that gradient computation hinges on all intermediate parameters within a network, we bring into play a strategy to not only make the gradients dissimilar but also to influence the intermediate layers of all networks within the ensemble. We aim to minimize the correlation between the outputs of the hidden layers, instigating enhanced diversity at each layer. This twofold approach of reducing correlation and enhancing diversity not only ensures different gradient paths but also fortifies the robustness of the ensemble to adversarial perturbations. Consequently, the  PARL framework presents a more potent defense against adversarial examples.

The pairwise similarity of gradients of the output of the $k^{th}$ hidden layer with respect to the parameters between $\mathcal{M}_{i}$ and $\mathcal{M}_{j}$ for a particular data point $\mathbf{x}$ can be represented as

\vspace{-\baselineskip} 
{\small
\begin{equation*}
\mathcal{G}_{k}^{(i, j)}(\mathbf{x}) = \cos\theta_{i,j}(\mathbf{x}) = \frac{\langle \nabla_{{w}^{k}_{i}}\mathcal{F}^{k}_{i}(\mathbf{x}), \nabla_{{w}^{k}_{j}}\mathcal{F}^{k}_{j}(\mathbf{x})\rangle}{|\nabla_{{w}^{k}_{i}}\mathcal{F}^{k}_{i}(\mathbf{x})| \cdot |\nabla_{{w}^{k}_{j}}\mathcal{F}^{k}_{j}(\mathbf{x})|}
\end{equation*}}
\vspace{-\baselineskip} 
\color{black}

\noindent where $\langle a, b \rangle$ represents the dot product between two vectors $a$ and $b$, and $\cos\theta_{i,j}(\mathbf{x})$ represents the cosine of the angle between two vectors. The overall pairwise similarity between $\mathcal{M}_{i}$ and $\mathcal{M}_{j}$ for $\mathbf{x}$ considering $\mathcal{H}$ hidden layers is given as $\mathcal{G}^{(i, j)}(\mathbf{x}) = \sum_{k=1}^{\mathcal{H}}\mathcal{G}_{k}^{(i, j)}(\mathbf{x})$

Additionally, we utilize intermediate outputs of the convolution layers to further diversify the decision boundaries. For this we define the pairwise similarity of outputs of the $k^{th}$ hidden layer with respect to input between $\mathcal{M}_{i}$ and $\mathcal{M}_{j}$ for a particular data point $\mathbf{x}$ as

\vspace{-\baselineskip} 
\begin{equation*}
\mathcal{L}_{k}^{(i, j)}(\mathbf{x}) = \rho(\mathcal{F}^{k}_{i}(\mathbf{x}),\mathcal{F}^{k}_{j}(\mathbf{x})) = \frac{\text{cov}(\mathcal{F}^{k}_{i}(\mathbf{x}), \mathcal{F}^{k}_{j}(\mathbf{x}))}{\sigma_{\mathcal{F}^{k}_{i}(\mathbf{x})} \sigma_{\mathcal{F}^{k}_{j}(\mathbf{x})}}
\end{equation*}
\vspace{-\baselineskip} 

where $\mathcal{L}_{k}^{(i, j)}(\mathbf{x})$ is the Pearson correlation between $\mathcal{F}^{k}_{i}(\mathbf{x})$ and $\mathcal{F}^{k}_{j}(\mathbf{x})$, $\rho$ is the Pearson correlation function, 
$\text{cov}(\mathcal{F}^{k}_{i}(\mathbf{x})$, $\mathcal{F}^{k}_{j}(\mathbf{x}))$ denotes the covariance between outputs of $\mathcal{F}^{k}_{i}(x)$ and $\mathcal{F}^{k}_{j}(x)$,
$\sigma_{\mathcal{F}^{k}_{i}(\mathbf{x})}$ and $\sigma_{\mathcal{F}^{k}_{j}(\mathbf{x})}$ denote the standard deviations of the sub-model outputs respectively. We define:
$\mathcal{L}^{(i, j)}(\mathbf{x}) = \sum_{k=1}^{\mathcal{H}}\mathcal{L}_{k}^{(i, j)}(\mathbf{x})$

\noindent Next, we define a penalty term $\mathcal{R}(\mathcal{M}_{i}, \mathcal{M}_{j})$ for all training examples in $\mathcal{X}$ to pairwise train $\mathcal{M}_{i}$ and $\mathcal{M}_{j}$ as

\vspace{-\baselineskip} 
{\small
\begin{equation*}
\mathcal{R}(\mathcal{M}_{i}, \mathcal{M}_{j}) = \frac{1}{|\mathcal{X}| \cdot |\mathcal{H}|}\sum_{\mathbf{x} \in \mathcal{X}} \left(\mathcal{G}^{(i, j)}(\mathbf{x}) \cdot \mathcal{L}^{(i, j)}(\mathbf{x}\right)
\end{equation*}}
\vspace{-\baselineskip} 

% \textcolor{blue}{
Here, the rationale behind multiplying the terms $\mathcal{G}$ and $\mathcal{L}$ lies in creating an interdependent relationship between the diversity of learning trajectories (as encouraged by gradient orthogonality) and the diversity of internal feature representations across the classifiers. By this design, a classifier's contribution to the ensemble's robustness is maximized only when it exhibits both gradient diversity and diverse feature representation simultaneously. This further ensures that the classifiers do not lean towards optimizing one aspect of diversity at the expense of the other.
% }

 We add $\mathcal{R}$ to training loss as a penalty parameter to penalize training for a large $\mathcal{R}$. \noindent $\mathcal{R}$ computes average pairwise similarity  for all training examples. $\mathcal{R}$ will gradually decrease as relative angles between the pair of gradients increases in higher dimension. Hence, the objective of PARL is to reduce $\mathcal{R}$. Thus, we add $\mathcal{R}$ to training loss as a penalty parameter to penalize training for a large $\mathcal{R}$.

In ensemble $\mathcal{M}_\mathcal{T}$, we compute $\mathcal{R}$ for each distinct pair of $\mathcal{M}_i$ and $\mathcal{M}_j$ in order to enforce diversity between each pair of classifiers. We define PARL to train~$\mathcal{M}_\mathcal{T}$ as

\vspace{-\baselineskip} 
{\small
\begin{equation}\label{eq:parl}
\begin{aligned}
PARL(\mathcal{M}_{\mathcal{T}}) &= \frac{1}{|\mathcal{X}|}\sum_{\mathbf{x} \in \mathcal{X}}\sum_{i=1}^{\mathcal{N}}J_{\mathcal{M}_{i}}(\mathbf{x}, \mathbf{y}) \ \\
&+ \gamma \cdot \sum_{1 \leq i < j \leq \mathcal{N}}\mathcal{R}(\mathcal{M}_{i}, \mathcal{M}_{j})
\end{aligned}
\end{equation}}
\vspace{-\baselineskip}

\noindent where $\gamma$ is a hyperparameter controlling the accuracy-robustness trade-off. A lower $\gamma$ enhances clean accuracy during ensemble training but reduces AE robustness. Conversely, a higher $\gamma$ boosts AE robustness while potentially sacrificing overall accuracy.

\color{black}

One may note that including a penalty for each distinct pair of classifiers within $\mathcal{M}_\mathcal{T}$ to compute PARL has one fundamental advantage. If we omit the pair $(\mathcal{M}_a, \mathcal{M}_b)$ in PARL computation, training will continue without any diversity restrictions between them. Consequently, producing similar decision boundaries that increase the likelihood of adversarial transferability between them, affecting the overall robustness of $\mathcal{M}_\mathcal{T}$. One may also note that in an efficient implementation of PARL one needs a single forward pass to get all the hidden layer outputs. Additionally, the number of gradient computations (backward pass) is directly proportional to the number of classifiers in $\mathcal{M}_\mathcal{T}$. The gradients for each classifier are computed once and are reused to compute $\mathcal{R}$ for each pair of classifiers. Reusing gradients protects the implementation from exponential computational overhead. Moreover, the complexity of gradient computation of PARL mostly depends on the architectural depth of neural networks. Given a fixed architecture, the training complexity of PARL grows linearly with the number of training examples. Hence, PARL is applicable to any dataset without adversely affecting the original training time.

\section{Experimental Evaluation}\label{sec:results}
\noindent \textbf{Evaluation Configurations: }
We consider two standard architectures Resnet20 and WideResnet28-10 for creating our ensembles. Each ensemble is a set of three sub-models of the same architecure\footnote{We select 3 sub-models to compare PARL with related methods, most of which use 3 sub-models. PARL is scalable for larger ensemble sizes.}. We consider CIFAR-10 and CIFAR-100, standard image classification datasets for our evaluation.  We consider four previously proposed countermeasures to compare the performance of PARL. We denote $ENS_{ADP}$, $ENS_{GAL}$, $ENS_{TRS}$, $ENS_{DVERGE}$, and $ENS_{EIO}$ to be the ensembles trained with the methods proposed in~\cite{DBLP:conf/icml/PangXDCZ19},~\cite{DBLP:journals/corr/abs-1901-09981},~\cite{DBLP:conf/nips/YangZDIGTWB020}, \cite{DBLP:conf/nips/YangLXZCZRZL21}, and ~\cite{DBLP:conf/aaai/0003NY023}, respectively. The ensemble trained with PARL is denoted as $ENS_{PARL}$. $ENS_U$ is the baseline ensemble model.

We use \textit{Adam optimization} to train all the ensembles with adaptive learning rate starting from $0.001$. We dynamically generate an augmented dataset using random shifts, flips and crops to train both CIFAR-10 and CIFAR-100.  We use $\gamma = 0.25$ as it provided best clean and robust accuracy trade-off (cf. Sec.~\ref{sec:ablation_study}), and \textit{categorical crossentropy} loss for $\mathcal{J}_{\mathcal{M}_i}(\cdot)$ for $ENS_{PARL}$ (ref. Equation~(\ref{eq:parl})). All ensembles are trained using two GPU servers: Intel Xeon CPU@2.30GHz with 16GB NVIDIA Tesla P100 GPU and Intel Xeon CPU@2.40GHz with 48GB NVIDIA A40.

We evaluate PARL, considering the same attacks as most recent defense EIO~\cite{DBLP:conf/aaai/0003NY023}. For black-box transfer attack, we use the following attacks: (1) PGD with momentum and three random starts~\cite{DBLP:conf/iclr/MadryMSTV18}; (2) M-DI$^2$-FGSM~\cite{DBLP:conf/cvpr/XieZZBWRY19}; and (3) SGM~\cite{DBLP:conf/iclr/Wu0X0M20}. The iterative steps are set to 100 with step size of $\epsilon$/5.
We use $\epsilon$ = \{0.01, 0.02, 0.03, 0.04, 0.05, 0.06, 0.07\} for generating AE of different strengths\footnote{We use majority voting as final ensemble decision for PARL. Hence, we use majority attack~\cite{DBLP:journals/corr/abs-2011-14031} for all adversarial attacks on PARL ensembles.}. We report the robust accuracy in all-or-nothing manner, meaning a sample is said to be correctly classified if all of its adversarial samples using different attack methods are correctly classified.

\begin{figure}[!t]
    \centering
    \subfloat[\label{fig:cka_cifar10_parl6}]{\includegraphics[width=1\linewidth]{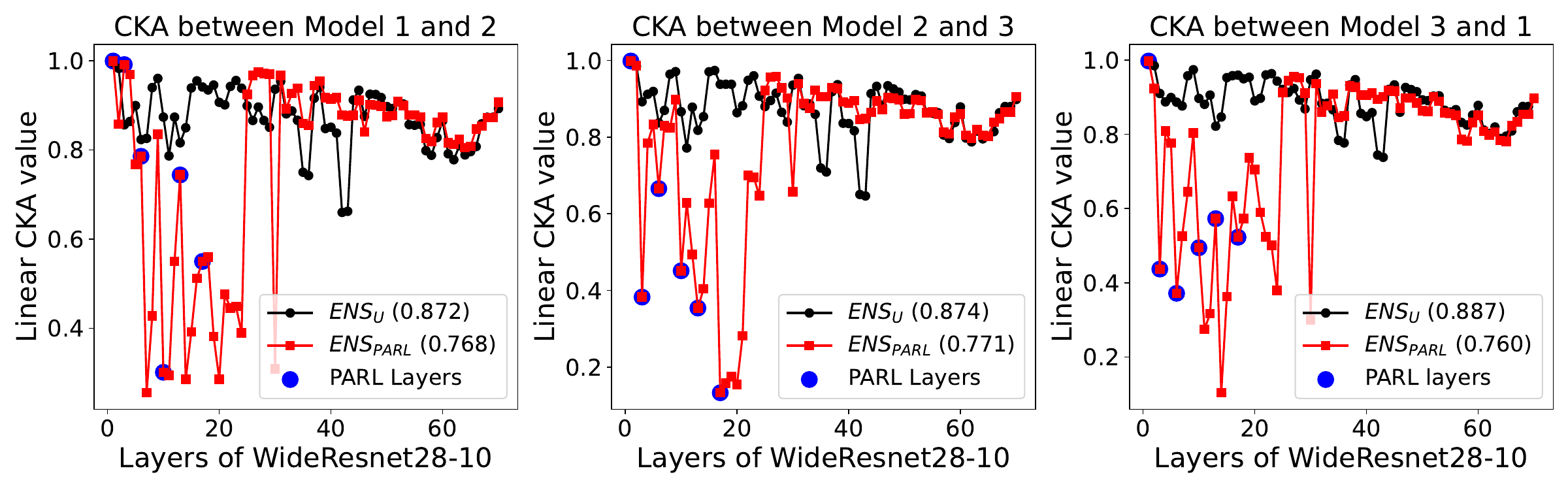}}
    % \vspace{-0.4cm}
    \quad
    \subfloat[\label{fig:cka_cifar10_parl3}]{\includegraphics[width=1\linewidth]{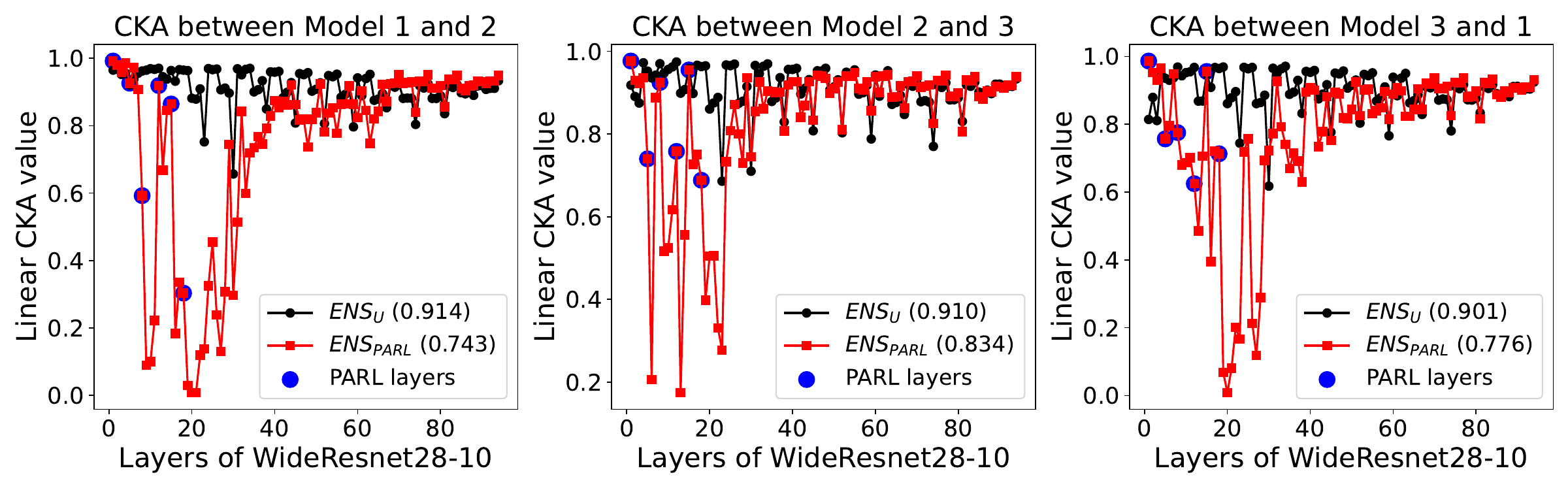}}
    \vspace{-0.3cm}
    \caption{\footnotesize Layer-wise linear CKA values between each pair of models trained with CIFAR-10 on (a) Resnet20 and (b) WideResnet28-10 showing the similarities at each layer.}  \vspace{-0.6cm}
    \label{fig:cka_cifar}
\end{figure}

\noindent \textbf{Analysing the Diversity: }
PARL aims to increase the diversity among all classifiers within an ensemble. To analyze the diversity of different classifiers trained using PARL, we use Linear Central Kernel Alignment (CKA) analysis~\cite{DBLP:conf/icml/Kornblith0LH19}. The CKA metric $\in [0, 1]$ measures similarity between decision boundaries represented by a pair of neural networks. A higher CKA indicates a significant similarity in decision boundary representations, which implies good transferability of AE. We present an analysis on layer-wise CKA for each pair of classifiers within $ENS_{\mathcal{U}}$ and $ENS_{PARL}$ trained with CIFAR-10 on Resnet20 and WideResnet28-10 architectures in Fig.~\ref{fig:cka_cifar} to show the effect of PARL on diversity.  We enforce diversity among all classifiers in $ENS_{PARL}$ for the first six convolution layers. We chose six layers as it provides better robust accuracy compared to other number of layers as shown in next subsection. We highlight the selected convolution layers for PARL in blue. There are intermediate layers between convolution layers as well such as batch-normalization and activation function.

We observe that each pair of models in $ENS_{\mathcal{U}}$ show a significant similarity at each layer. However, since $ENS_{PARL}$ restricts the first six convolution layers, we observe a notable decline in CKA values at initial layers. The observation is expected as PARL  imposes layer-wise diversity in its formulation. The overall average Linear CKA values between each pair of models in Fig.~\ref{fig:cka_cifar} are mentioned inside brackets within corresponding figure legends, which signifies that the classifiers within an ensemble trained using PARL shows a higher overall dissimilarity than the unprotected baseline ensemble. Next, we analyze the effect of observed diversity on the performance of $ENS_{PARL}$.

\vspace{-0.2cm}
\subsection{Robustness Evaluation}
\begin{figure}[!t]
    \centering
    \subfloat[\label{fig:ens_cifar10_bb}]{\includegraphics[width=0.99\linewidth]{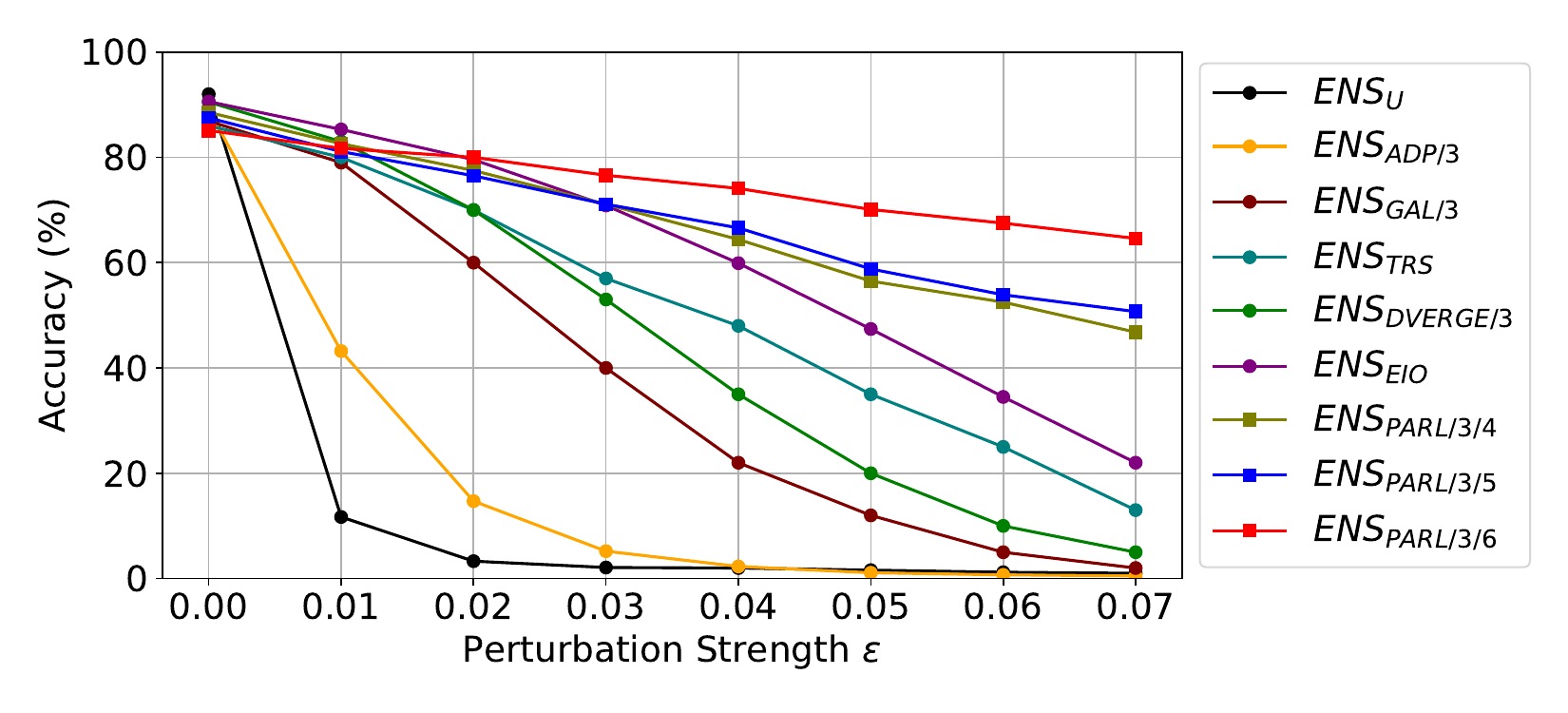}}
    % \vspace{-0.4cm}
    \quad
    \subfloat[\label{fig:ens_cifar100_bb}]{\includegraphics[width=0.99\linewidth]{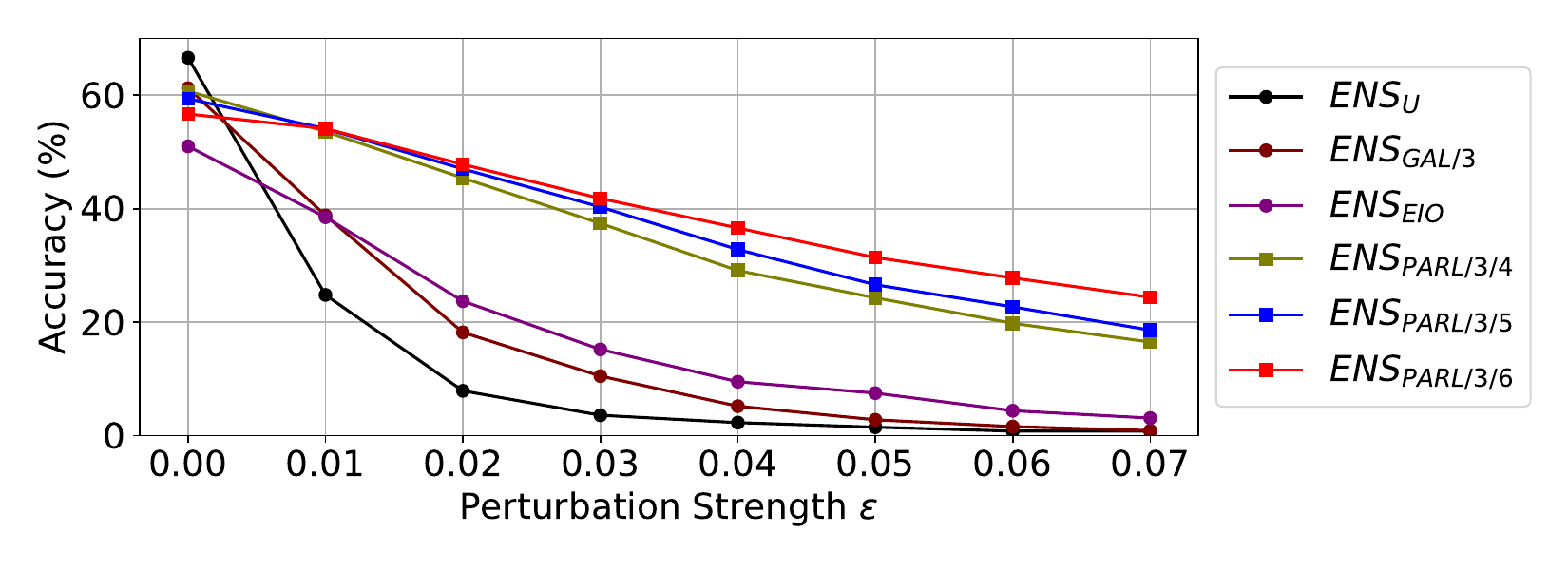}}
    \vspace{-0.3cm}
    \caption{\footnotesize Resnet20 Ensemble classification accuracy (\%) vs. Attack Strength ($\epsilon$) against black-box transfer attacks generated from surrogate ensemble with (a) CIFAR-10 and (b) CIFAR-100 dataset  \vspace{-0.7cm}}
    \label{fig:ens_cifar10}
\end{figure}

\vspace{-0.2cm}
The attacker cannot access the model parameters and rely on surrogate models to generate transferable AE. Under such a black-box scenario, we use one hold-out ensemble with three Resnet20 architectures as the surrogate model. We randomly select 1000 test samples and evaluate the performance of black-box transfer attacks for all ensembles across a wide range of attack strength $\epsilon$. 
 We give a detailed performance evaluation considering multiple attack strengths for CIFAR-10 and CIFAR-100 dataset in Fig.~\ref{fig:ens_cifar10_bb} and Fig.~\ref{fig:ens_cifar100_bb} respectively. To avoid confusion with nomenclature of other ensemble defenses, $ENS_{PARL/3/4}$, $ENS_{PARL/3/5}$ and  $ENS_{PARL/3/6}$ indicates an ensemble of 3 classifiers with 4, 5 and 6 initial convolution layers modified with the PARL loss respectively. On the CIFAR-10 dataset, we note some key observations. The model $ENS_{PARL/3/6}$  with a clean accuracy of 85.09\%  performs the best among all the previous ensemble defense methods. It is to be noted, for $\epsilon = 0.07$, $ENS_{PARL/3/6}$ has a robust accuracy of $64.6\%$, which is $42.6\%$ higher than the previous state-of-the-art defense $ENS_{EIO}$, but has drop of $5\%$ clean accuracy. On the other hand $ENS_{PARL/3/5}$ and $ENS_{PARL/3/4}$ show an increase of $28.7\%$ and $24.8\%$ robust accuracy for $\epsilon = 0.07$ with just $3.3\%$ and $2\%$ drop in clean accuracy compare to $ENS_{EIO}$. For the CIFAR-100 dataset, PARL surpasses state-of-the-art method both in terms of robust as well as clean accuracy. These results suggest that, based on the desired robustness level, a model architect can adjust the number of layers influenced by the PARL loss function. This fine-tuning enables achieving the desired defense against adversarial attacks, with only minor trade-offs in clean accuracy. PARL primarily focuses on defending against black-box transfer attacks. These adversaries can craft adversarial examples using the complete network parameters, although such scenarios are rarely practical in real-world applications where only API based query access is provided for the target model.

\noindent \textbf{Performance evaluation against Query-based black-box attack:}
In a query-based adversarial attack, the attacker crafts adversarial examples by analyzing the responses from the target machine learning model, to which they have only black-box access. Our assessment of PARL involves its performance against a specific query-based black-box adversarial attack known as the Square Attack~\cite{DBLP:conf/eccv/AndriushchenkoC20}). This attack method is notable for its ability to efficiently modify a minimal number of pixels within a square area of an image, effectively deceiving machine learning models while maintaining the overall visual integrity of the image. The Square Attack is unique in the AutoAttack suite~\cite{DBLP:conf/icml/Croce020a}, which comprises four different methods of adversarial attacks used for thorough model evaluation, with Square Attack being the sole black-box method.

\begin{figure}[!t]
    \centering
    \subfloat[\label{fig:ens_square}]{\includegraphics[width=0.99\linewidth]{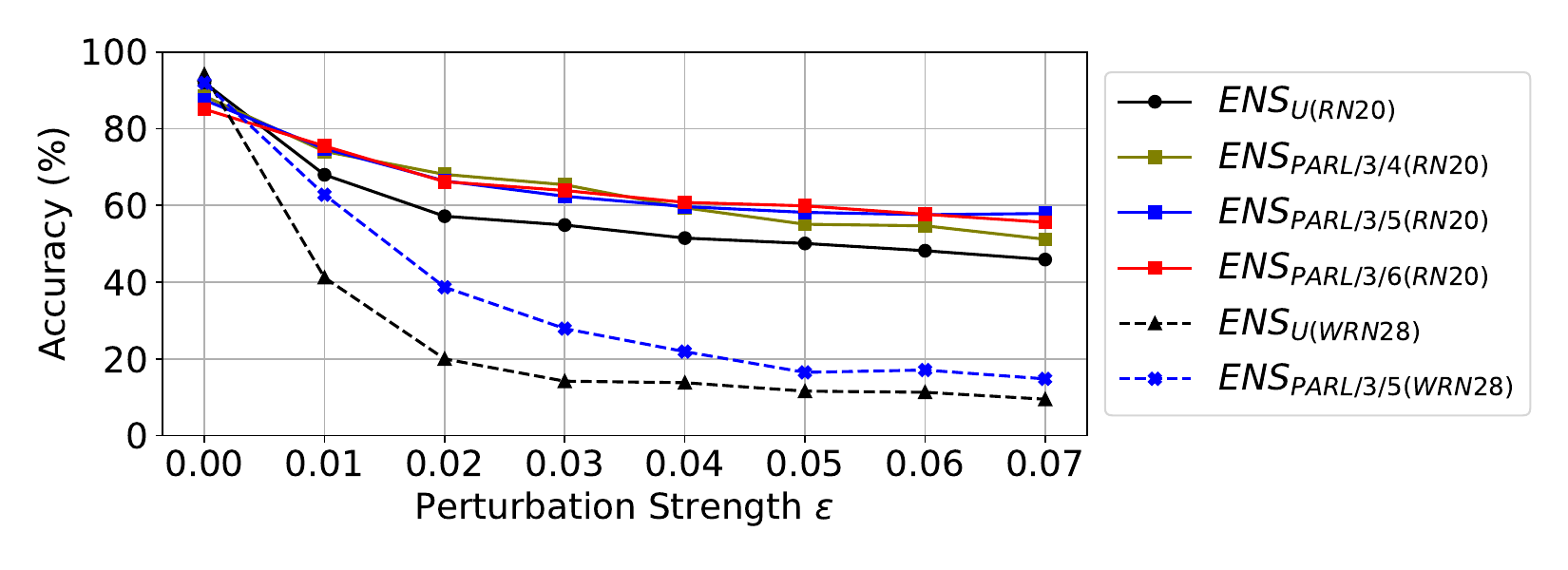}}
    % \vspace{-0.4cm}
    \quad
    \subfloat[\label{fig:ens_square_query}]{\includegraphics[width=0.99\linewidth]{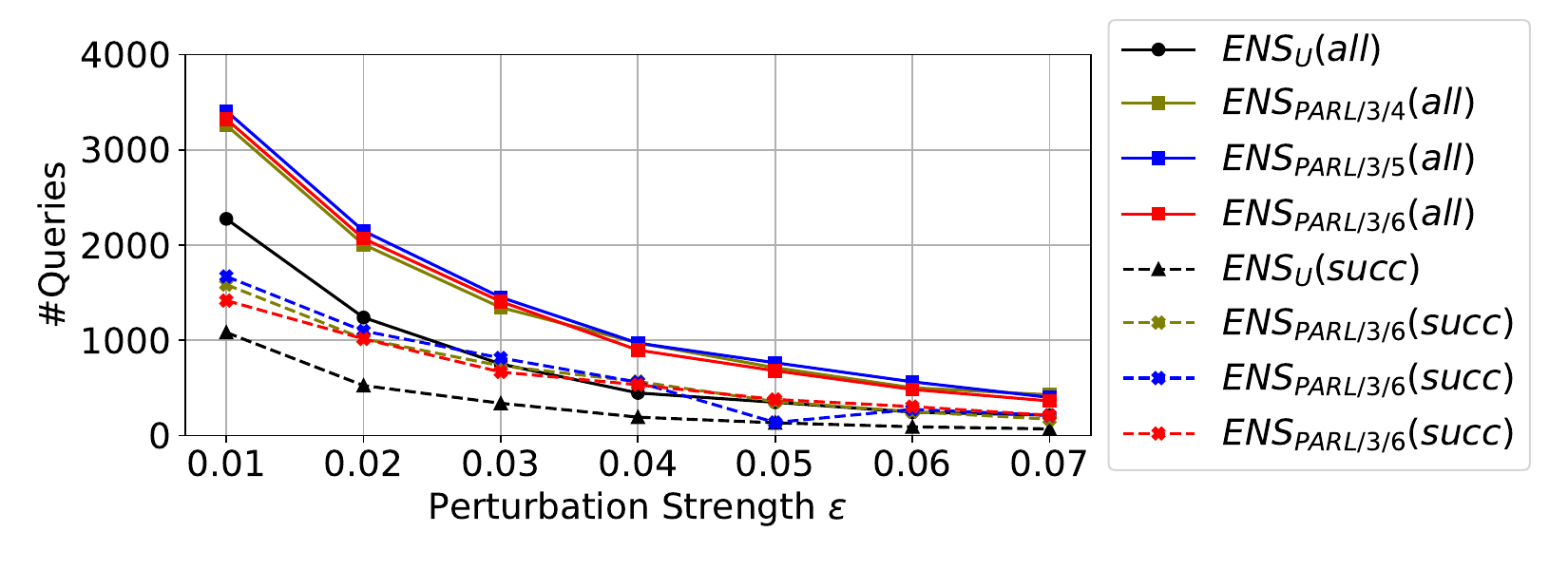}}
    \vspace{-0.3cm}
    \caption{\footnotesize (a) Resnet20 (RN20) and WideResnet28-10 (WRN28-10) Ensemble classification accuracy (\%) vs. Attack Strength ($\epsilon$) against Square attack for CIFAR-10 (b) Average number of queries required for square attack for all samples as well as successful (succ) attack samples with RN20 \vspace{-0.7cm}}
    \label{fig:square}
\end{figure}

In Fig.~\ref{fig:ens_square}, we illustrate the outcomes of using the Square Attack on PARL. For this experiment, we used the default maximum queries and square size settings, which are $5000$ and $0.8$, respectively and considered $1000$ test images. We found that ensembles of Resnet20 and WideResnet28-10 trained with PARL outperformed the surrogate model. Additionally, we calculated the average number of queries utilized across all test images and specifically for those images which were successfully misclassified by the ensemble. Our findings, displayed in Fig.~\ref{fig:ens_square_query}, indicate that all models trained with PARL required over twice the number of queries compared to the surrogate model for all levels of perturbation, except at $0.01$, where the requirement was approximately 1.5 times higher. PARL's ability to require significantly more queries for successful adversarial attacks, especially in comparison to the surrogate model, demonstrates its robustness in less query-restrictive environments.

\begin{figure}[!t]
    \centering
    \subfloat[\label{fig:ens_cifar10_bb_trades}]{\includegraphics[width=0.8\linewidth]{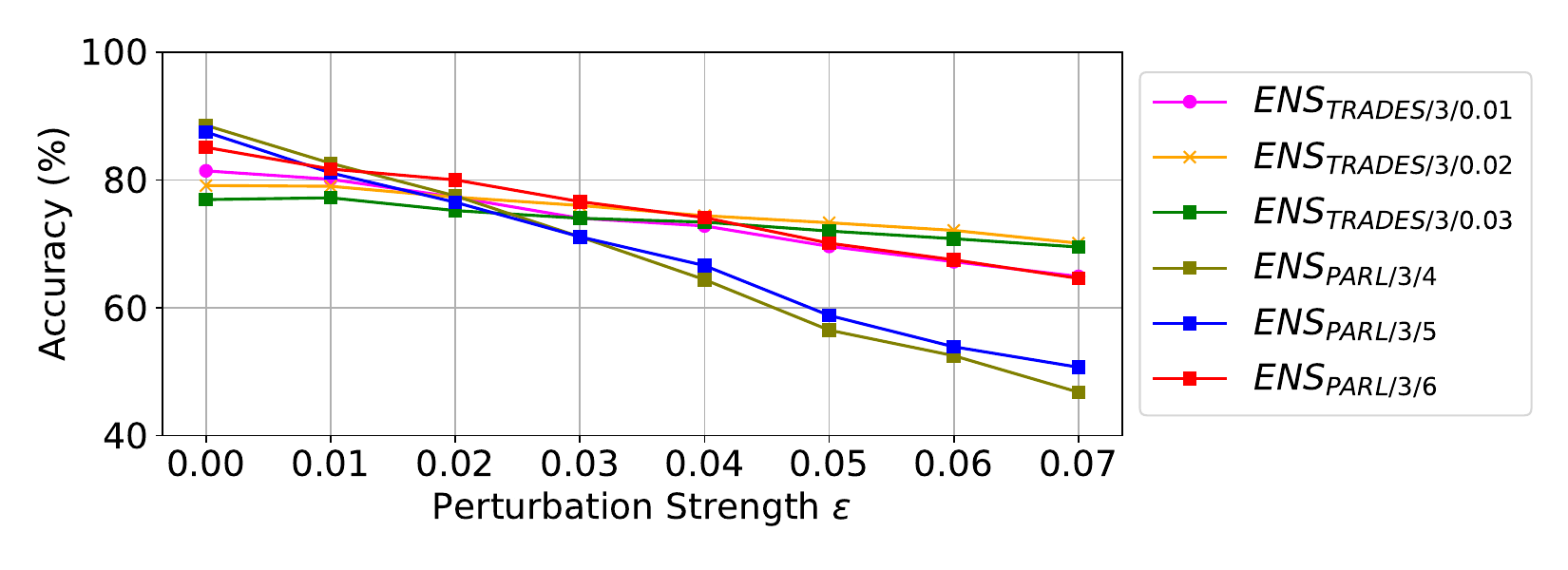}}
    % \vspace{-0.49cm}
    \quad
    \subfloat[\label{fig:ens_cifar100_bb_trades}]{\includegraphics[width=0.8\linewidth]{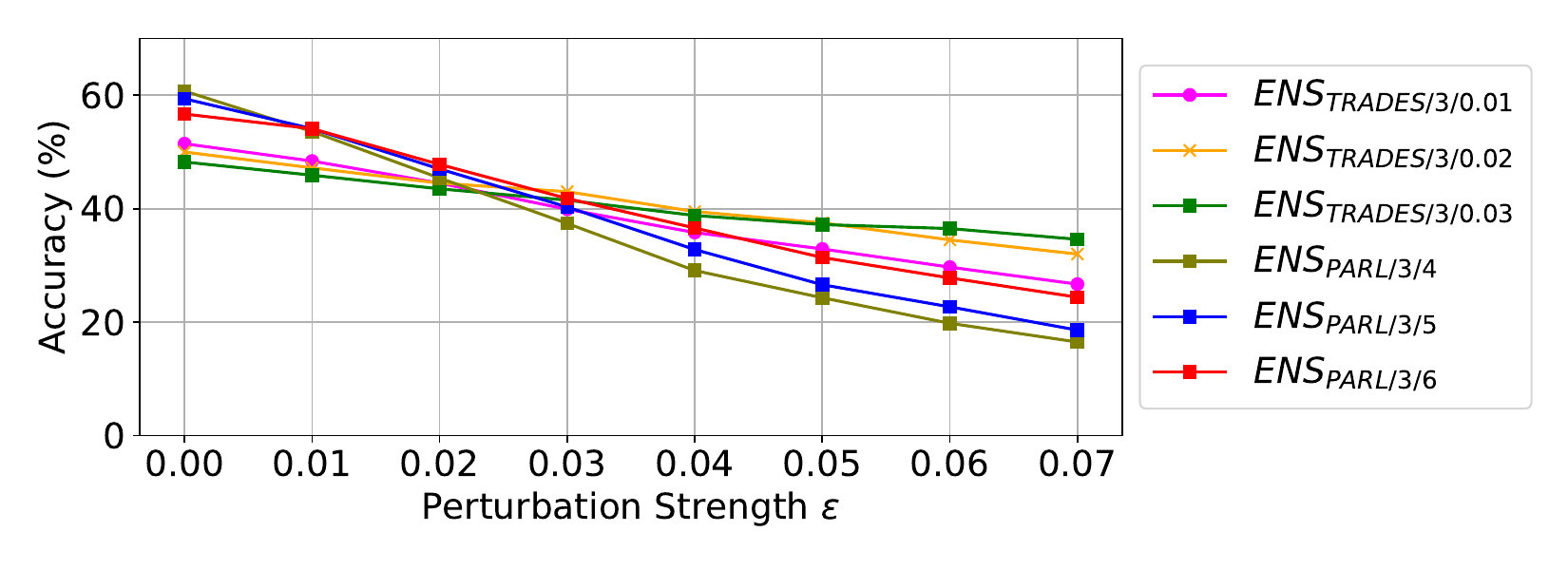}}
    \vspace{-0.3cm}
    \caption{\footnotesize Resnet20 Ensemble classification accuracy (\%) vs. Attack Strength ($\epsilon$) against black-box transfer attacks generated from surrogate ensemble for (a) CIFAR-10 and (b) CIFAR-100 \vspace{-0.65cm} }
    \label{fig:trades}
\end{figure}

\noindent \textbf{Performance comparison with Adversarial Training:}
We evaluate the performance of the PARL model in comparison with the adversarial training method TRADES~\cite{DBLP:conf/icml/ZhangYJXGJ19}. The TRADES loss function is defined as:

% \vspace{-\baselineskip} 
% \begin{equation}
% \mathcal{L}_{\text{TRADES}}(x, y) = \mathcal{L}(x, y) + \beta \cdot \mathcal{L}(x + \delta, y)
% \end{equation}
% \vspace{-\baselineskip} 

\vspace{-5mm} % Adjust as needed
\begin{equation}
\mathcal{L}_{\text{TRADES}}(x, y) = \mathcal{L}(x, y) + \beta \cdot \mathcal{L}(x + \delta, y)
\end{equation}
\vspace{-\baselineskip}

\noindent where, \( x \) is the natural input and \( y \) is its corresponding label. \( \mathcal{L}(x, y) \) is the natural loss representing the model's prediction error on the clean data. \( \delta \) represents the adversarial perturbation, typically computed using methods like PGD. \( \mathcal{L}(x + \delta, y) \) is the adversarial loss, emphasizing correct classification of adversarial examples. \( \beta \) is a hyperparameter that balances the contributions of the two terms. This approach differs from standard adversarial training which often seeks to minimize the adversarial loss \( \mathcal{L}(x + \delta, y) \) alone. By combining both the natural and adversarial losses, TRADES ensures robustness against adversarial attacks while maintaining performance on clean examples. We assume the default  \( \beta \) value of 6 for our experiments. In Fig.  \ref{fig:ens_cifar10_bb_trades} and Fig. \ref{fig:ens_cifar100_bb_trades} we show the comparison of three adversarially trained TRADES Resnet20 ensemble models $ENS_{TRADES/3/0.01}$,  $ENS_{TRADES/3/0.02}$ and $ENS_{TRADES/3/0.03}$ (trained with different PGD attack perturbations $\epsilon = 0.01, 0.02, 0.03$) against $ENS_{PARL/3/4}$, $ENS_{PARL/3/5}$ and $ENS_{PARL/3/6}$ for CIFAR-10 and CIFAR-100 respectively. For CIFAR-10, we observe that at $\epsilon=0.07$ $ENS_{PARL/3/6}$ gives only $5.5\%$ less robust accuracy than  $ENS_{TRADES/3/0.02}$ and $ENS_{TRADES/3/0.03}$ with clean accuracy $5.97\%$ and $8.17\%$ higher than them respectively. Additionally, it gives same robust accuracy as $ENS_{TRADES/3/0.01}$ with $3.68\%$ higher clean accuracy. We observe similar trends in the results for the CIFAR-100 dataset as well. In conclusion, PARL stands out as a preferred defense strategy against adversarial attacks on ensembles, offering similar or slightly lower robust accuracy compared to TRADES but with significantly higher clean accuracy, all achieved in less than one-third of TRADES's training time (cf. Sec.~\ref{sec:ablation_study}).

\begin{table}[h]
\vspace{-0.3cm}
\caption{\footnotesize Resnet20 Ensemble clean and robust ($\epsilon = 0.01$) classification accuracy on Tiny Imagenet Dataset \vspace{-0.3cm}}
\label{res_tinyimagenet}
\centering
\scriptsize
\begin{tabular}{ccc}
\hline
\textbf{Model} & \textbf{Clean Accuracy} & \textbf{Robust Accuracy} \\ \hline
\textbf{$ENS_{U}$} & 60.85\% & 17.8\% \\ \hline
\textbf{$ENS_{EIO}$} & 57.33\% & 8.6\% \\ \hline
\textbf{$ENS_{TRADES/3/0.01}$} & 44.32\% & 27.7\% \\ \hline
\textbf{$ENS_{PARL/3/4}$} & 55.95\% & 28.3\% \\ \hline
\textbf{$ENS_{PARL+TRADES}$} & 42\% & 35.6\% \\ \hline
\end{tabular}
\vspace{-0.4cm}
\end{table}

\noindent \textbf{Performance evaluation on Tiny Imagenet Dataset: }
We evaluated PARL on the Tiny Imagenet dataset, which contains 200 classes. Table~\ref{res_tinyimagenet} presents the clean and robust accuracy results for EIO, TRADES, PARL, and a combination of PARL and TRADES. The state-of-the-art method $ENS_{EIO}$ underperformed, with both robust and clean accuracy falling below the baseline ensemble\footnote{We used the open-source code provided with the EIO paper and reported all results based on runs using the default configuration described in the paper.}. While $ENS_{TRADES/3/0.01}$ and $ENS_{PARL/3/4}$ showed similar robust accuracy, $ENS_{PARL/3/4}$ achieved better clean accuracy. Additionally, we also combined PARL and TRADES losses and observed an improvement in robust accuracy by nearly 7\%, but clean accuracy dropped to 42\%. Hence overall, in terms of clean accuracy and robust accuracy trade-off PARL performs the best among all methods. Overall, PARL offers the best trade-off between clean and robust accuracy among all methods tested.

% \footnote{We used the open-sourced code from the EIO paper. While our reproduced results for CIFAR10 were similar to those reported in the paper, the results for Tiny Imagenet did not match. Despite reaching out to the authors for guidance, we received no response. Therefore, we present the results we achieved for comparison.}
\vspace{-0.2cm}
\subsection{Ablation Study}
\label{sec:ablation_study}
\begin{figure}[h]
\centering
    \includegraphics[width=\columnwidth]{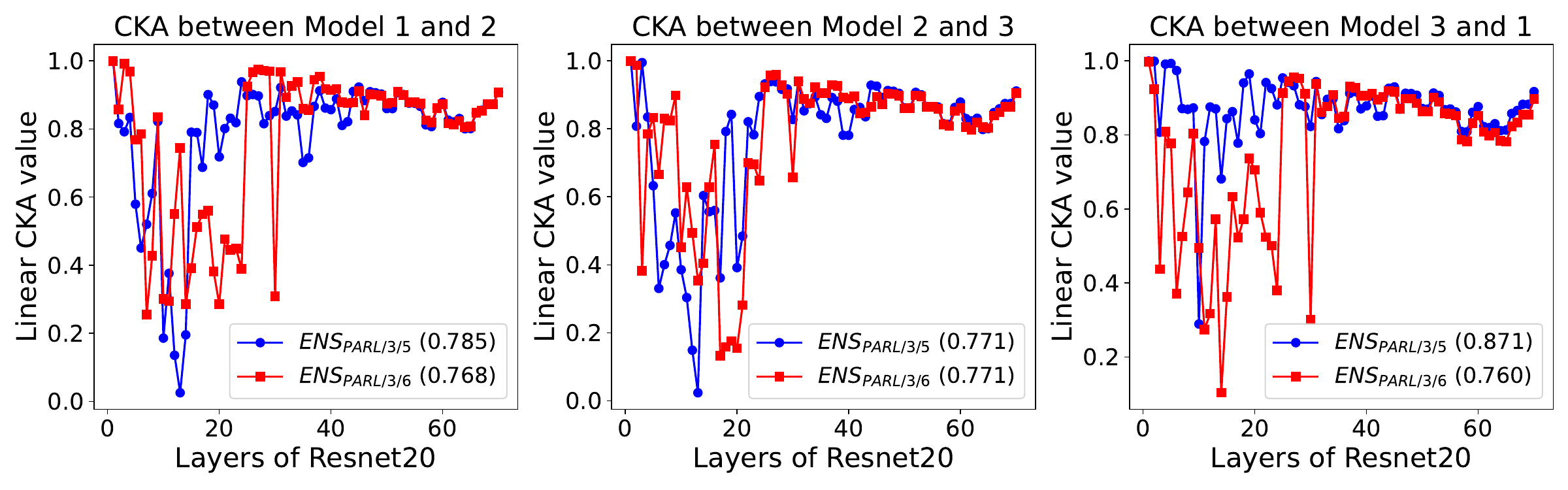}
    \vspace{-0.7cm}
    \caption{\footnotesize Layer-wise linear CKA values between each pair of PARL/3/5 and PARL/3/6 models trained with CIFAR-10 on Resnet20 showing the diversity at each layer. \vspace{-0.3cm}}
    \label{fig:ablation_cka_parl}
\end{figure}

\vspace{-0.2cm}
In our previous evaluations, we train $ENS_{PARL}$ by enforcing diversity in the first four, five and six convolution layers for all classifiers. Next, we provide an ablation study by analyzing a varying number of convolution layers considered for diversity training. We consider three ensembles, $ENS_{PARL/3/4}$, $ENS_{PARL/3/5}$, and $ENS_{PARL/3/6}$, for this study. Accuracies of all ensembles on clean examples for  Resnet20 and WideResnet28-10  are mentioned in Table~\ref{table:ablation_accuracy}. We observe that as fewer restrictions are imposed, overall ensemble accuracy increases, which is expected and can be followed from Equation~(\ref{eq:parl}). We also present a layer-wise CKA analysis for each pair of classifiers within $ENS_{PARL/3/5}$ and $ENS_{PARL/3/6}$, trained with CIFAR-10. The layer-wise CKA values are shown in Fig.~\ref{fig:ablation_cka_parl} to exhibit the effect of PARL on diversity. We observe a decline in the CKA values for more layers in case of $ENS_{PARL/3/6}$ compared to $ENS_{PARL/3/5}$, which is expected as $ENS_{PARL/3/6}$ is trained by restricting more convolution layers. We also observe that each pair of classifiers show more overall diversity in $ENS_{PARL/3/6}$ than in  $ENS_{PARL/3/5}$. The overall average of CKA values are mentioned inside braces within figure legends.

\begin{table}[h]
\vspace{-0.3cm}
\caption{\footnotesize Ensemble classification accuracy (\%) for Resnet20 and WideResnet28-10 on CIFAR-10 and CIFAR-100 clean examples. \vspace{-0.5cm}\label{table:ablation_accuracy}}
\begin{center}
\resizebox{0.9\linewidth}{!}{%
\begin{tabular}{lcccc}
\toprule
Model & Dataset & $ENS_{PARL/3/4}$ & $ENS_{PARL/3/5}$ & $ENS_{PARL/3/6}$ \\ 
\midrule
\multirow{2}{*}{Resnet20} & CIFAR-10 & 88.53 & 87.49 & 85.79  \\
                          & CIFAR-100 & 60.73 & 59.37 & 56.68 \\
\midrule
WideResnet28-10  & CIFAR-10 & 92.86 & 92.31 & 91.98 \\
\bottomrule
\end{tabular}}
\vspace{-0.9cm}
\end{center}
\end{table}

\begin{figure}[!t]
    \centering
    \subfloat[\label{fig:ablation_gradonly}]{\includegraphics[width=0.8\linewidth]{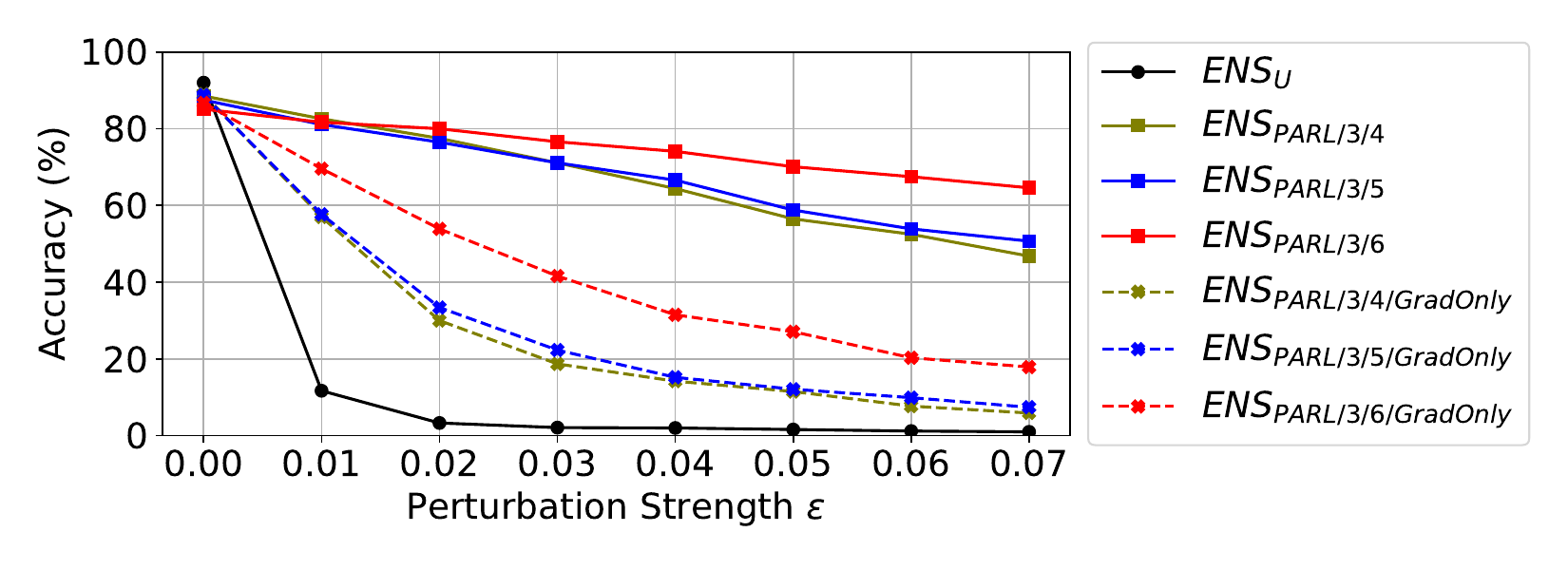}}
    % \vspace{-0.4cm}
    \quad
    \subfloat[\label{fig:ablation_gamma}]{\includegraphics[width=0.8\linewidth]{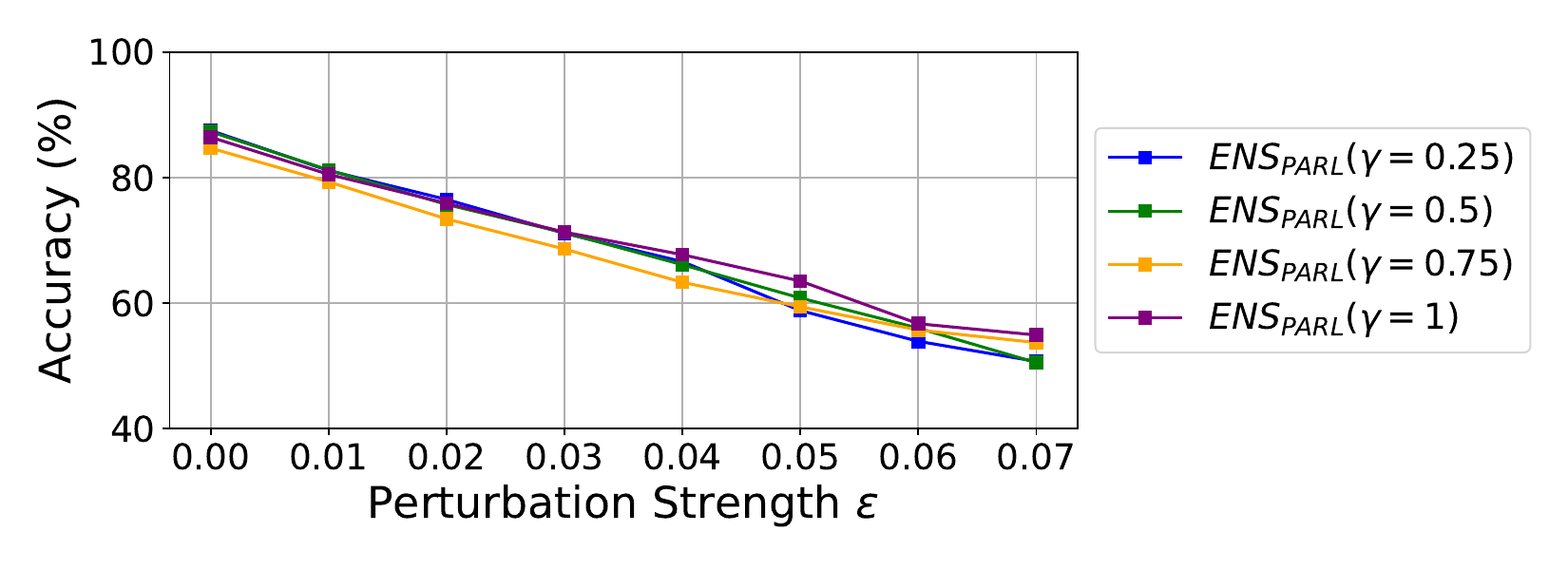}}
    \vspace{-0.3cm}
    \caption{\footnotesize (a) Comparing PARL robustness against PARL loss with penalty term that only uses gradient similarity. (b) Comparing clean and robust accuracies of Resnet20 PARL ensemble trained with different $\gamma$ \vspace{-0.5cm}}
    \label{fig:ablation}
\end{figure}

\noindent\textbf{Contribution of correlation term: }In the defined PARL loss function (see Equation \ref{eq:parl}), we incorporate a penalty term that combines the cosine similarities of gradients with the correlation of outputs from distinct sub-model pairs at specific convolution layers. 
Fig.~\ref{fig:ablation_gradonly} illustrates the variance in PARL's robustness when the penalty term solely relies on the cosine similarities between gradients, excluding the output correlations ($ENS_{PARL/3/N/GradOnly}$). While $ENS_{PARL/3/N/GradOnly}$ models do show improvements over $ENS_{U}$, it's evident that $ENS_{PARL/3/N}$ models are superior, emphasizing the critical role of both gradient similarity and output correlation in the penalty term.

\noindent\textbf{Selection of $\gamma$:} In PARL Equation \ref{eq:parl}, $\gamma$ is used for the purpose of regulating the PARL penalty term. We experimented with varying $\gamma$ values, focusing on their effect on the model's clean and robust accuracies. The findings for Resnet-20 $ENS_{PARL/3/5}$, illustrated in Fig. ~\ref{fig:ablation_gamma}, reveal that while accuracy fluctuations are minimal across different perturbations, lower $\gamma$ values tend to enhance clean accuracy, whereas higher $\gamma$ values improve robust accuracy. For $\gamma = 1$, we obtain a clean accuracy of 86.42\% and robust accuracy of 54.9\% at $\epsilon = 0.07$, whereas we for $\gamma = 0.25$ we obtain increased clean accuracy of 87.49\%, and decreased robust accuracy of 50.7\%. For our experiments, we selected $\gamma = 0.25$ as it offers the best trade-off between clean and robust accuracy.

\begin{figure}[!t]
\centering
    \includegraphics[width=0.8\columnwidth]{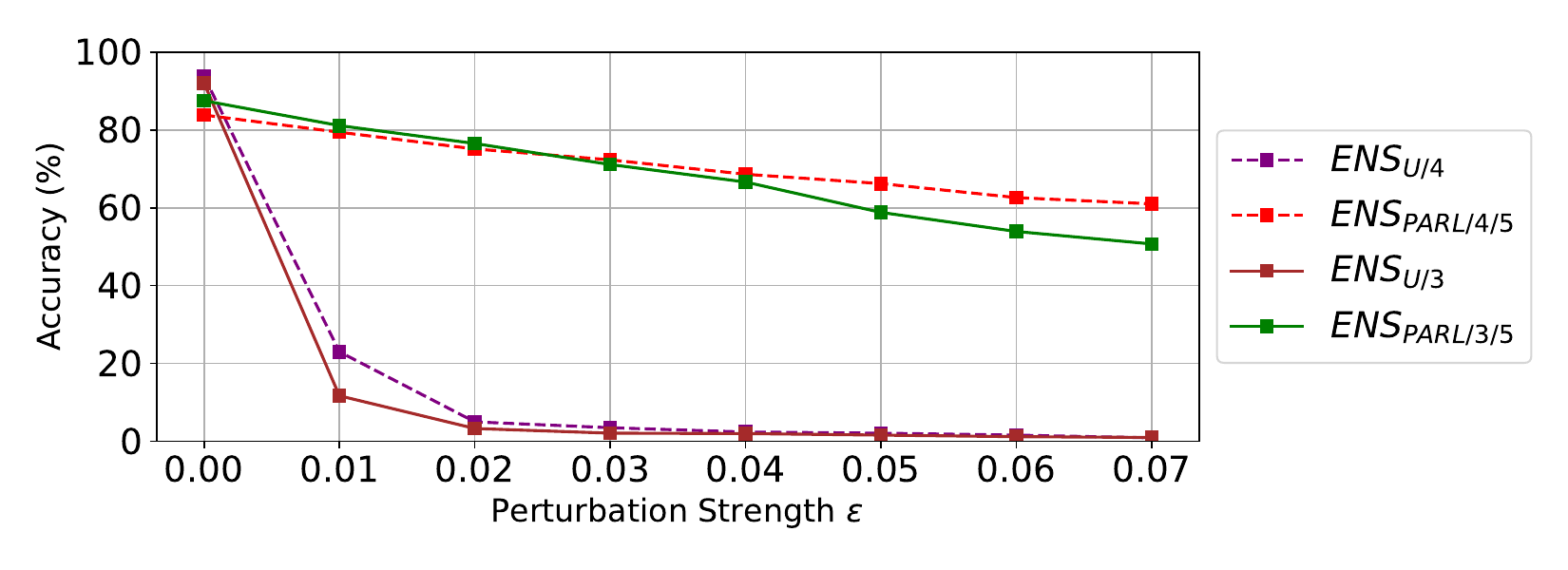}
    \vspace{-0.3cm}
    \caption{\footnotesize Resnet20 Ensemble classification accuracy (\%) vs. Attack Strength ($\epsilon$) for CIFAR-10 with increased number or classifiers \vspace{-0.7cm}}
    \label{fig:ens_3_4}
\end{figure}

\noindent\textbf{Increased number of classifiers: }
In Fig.~\ref{fig:ens_3_4}, we present results obtained by increasing the number of classifiers from three to four. We observe that robust accuracy improves by 10.3\% for $ENS_{PARL/4/5}$ compared to $ENS_{PARL/3/5}$  with $\epsilon = 0.07$, with 28.8\% more training time, as discussed next. We opt for three classifiers throughout the paper as a trade-off between robust accuracy and train time.

\noindent\textbf{Training time overhead:} In Table~\ref{table:ablation_time_accuracy} we give training time per epoch for the Resnet20 surrogate, PARL, EIO and TRADES models.  Earlier we compared PARL with TRADES and observed that PARL gives similar or slightly lesser robust accuracy than TRADES but in trade-off provides much higher clean accuracy as well. TRADES also takes 3x more training time than the most computationally expensive $ENS_{PARL/3/6}$ among all PARL models and 12x compared to the surrogate model. Additionally, $ENS_{EIO}$ takes 7x training time compared to surrogate model and 3x compared to $ENS_{PARL/3/4}$ which gives similar clean accuracy and much higher robust accuracy compared to $ENS_{EIO}$. Lastly, we observe that increasing the number of classifiers from three to four has minimal impact on training time, as shown for $ENS_{PARL/4/5}$ and $ENS_{PARL/4/6}$.

\begin{table}[!t]
\centering
\scriptsize
\caption{\footnotesize Training time (sec/epoch) for CIFAR-10. \vspace{-0.2cm}}
\label{table:ablation_time_accuracy}
\begin{tabular}{ccccc}
\toprule
 Model & Training Time & Model & Training Time \\
\midrule
$ENS_{U}$ & 30 & $ENS_{PARL/3/5}$ & 90 \\
$ENS_{EIO}$ & 210 & $ENS_{PARL/3/6}$ & 120 \\
$ENS_{TRADES}$ & 370 & $ENS_{PARL/4/5}$ & 116 \\
$ENS_{PARL/3/4}$  & 65 &  & \\
\bottomrule
\end{tabular}
\vspace{-0.6cm}
\end{table}

\noindent \textbf{Discussion:} ADP forces different models in an ensemble to have mutually orthogonal non-maximal predictions. GAL reduces the dimension of adversarial sub-space shared between different models using uncorrelated loss functions. These methods do not inherently enforce diversity on decision boundaries learned by the models. DVERGE diversifies non-robust input features of models by performing adversarial training, making it more robust against weak attack strength and less robust against strong attack strength. EIO leverages random gated networks to enhance adversarial robustness by diversifying vulnerabilities across multiple paths of CNNs but again has a higher training overhead. In contrast, PARL, by construction, forces the models to have high diversity in decision boundaries using all intermediate feature space. The diversity of models attained through intermediate feature space (not limited to only non-robust input features) makes PARL more robust, even for strong attack strength. In addition, PARL produces robust ensembles without substantially impacting clean example accuracy and training time.

\section{Conclusion}
\vspace{-0.25cm}
This paper proposes a new approach that, by construction, produces an ensemble of neural networks with diverse decision boundaries, making it robust against adversarial attacks. The diversity is obtained through the proposed Pairwise Adversarially Robust Loss (PARL) function utilizing the gradients and outputs of each layer in all the networks simultaneously. Experimental results show that PARL can significantly improve the overall robustness of an ensemble in comparison to previous approaches against state-of-the-art black-box transfer attacks as well as query-based black-box attacks without substantially impacting clean example accuracy. In particular, PARL achieves a $24.8\%$ improvement in robust accuracy over the leading ensemble defense method EIO with highest perturbation strength. Furthermore, when compared to TRADES, PARL demonstrates robust accuracy of similar order with a 3.68\% increase in clean accuracy. PARL also takes lesser training time compared to both EIO and TRADES method.

\section*{Acknowledgment}
\vspace{-0.3cm}
\noindent This work was supported in part by Prime Minister's Research Fellowship, granted by Government of India.

%%%%%%%%% REFERENCES
{\small
\bibliographystyle{ieee_fullname}
\bibliography{egbib}
}

\appendix

\section{White-box Adversary Evaluation}
\label{sec:appendix}

\begin{figure}[h]
\centering
    \includegraphics[width=\columnwidth]{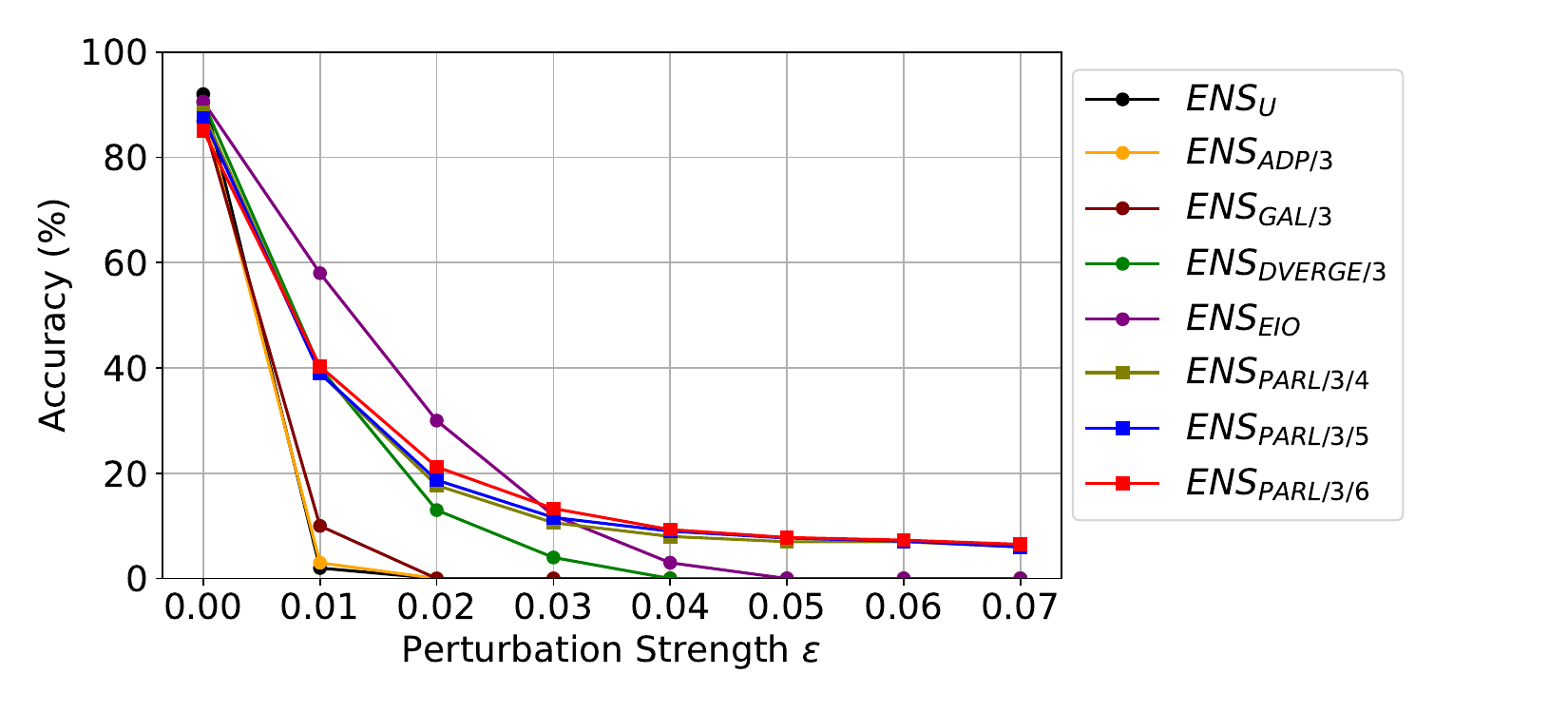}
    \caption{\footnotesize Resnet20 Ensemble classification accuracy (\%) vs. Attack Strength ($\epsilon$) against  white-box attacks for CIFAR-10 \vspace{-0.4cm}}
    \label{fig:ens_cifar10_wb}
\end{figure}

% \noindent \textbf{Performance considering Adversary with white-box access: }
The attacker has complete access to the model parameters. Under such a white-box scenario, we craft AE from the target ensemble itself. We randomly select 1000 test samples and evaluate white-box attacks for all ensembles across a wide range of attack strength $\epsilon$. We present the results for CIFAR-10 with Resnet20 model in Fig.~\ref{fig:ens_cifar10_wb}.  We observe that for lower perturbations $ENS_{PARL}$  performs similar to DVERGE whereas from $0.03$ onwards PARL performs better than the previous defenses. Though PARL's robustness against white-box attacks is still quite low, and it is a limitation which we plan to improve in our future works.

\section{Evaluation on VGG 16 and LeNet-5}

\begin{figure}[h]
\centering
    \begin{subfigure}[b]{0.49\columnwidth}
        \centering
        \includegraphics[width=\textwidth]{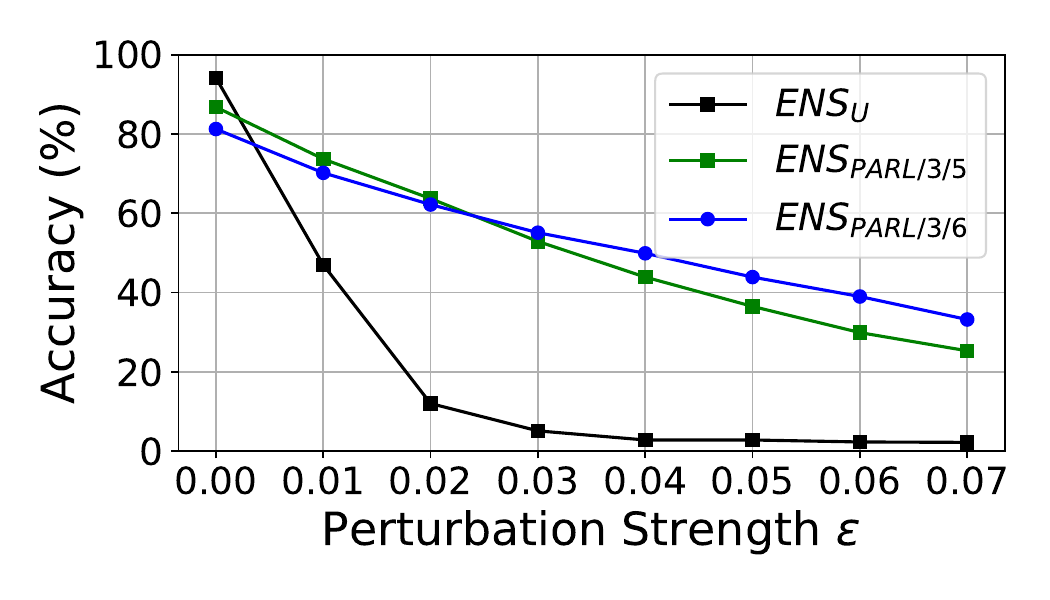}
        \caption{\footnotesize VGG16}
        \label{fig:ens_vgg16}
    \end{subfigure}%
    % \hfill
    \begin{subfigure}[b]{0.49\columnwidth}
        \centering
        \includegraphics[width=\textwidth]{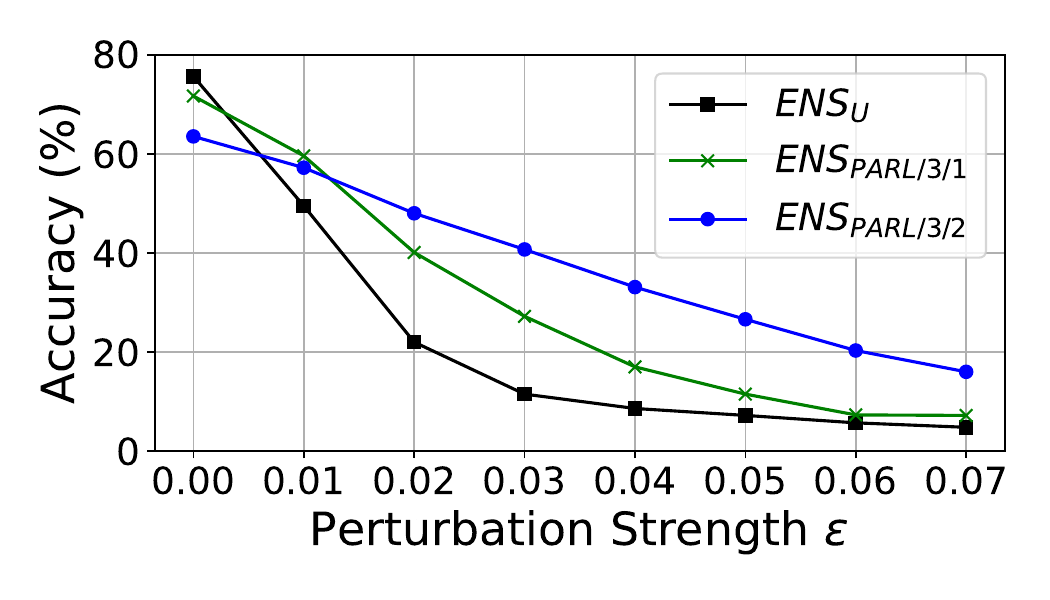}
        \caption{\footnotesize LeNet-5}
        \label{fig:ens_lenet}
    \end{subfigure}
    \caption{Ensemble classification accuracy (\%) vs. Attack Strength ($\epsilon$) for CIFAR-10 with different architectures \vspace{-0.4cm}}
    \label{fig:ens_combined}
\end{figure}

In the main paper, we presented results for PARL using ResNet models. To showcase its generalizability across other standard CNN architectures, including VGG16 and smaller models like LeNet-5 with only two convolutional layers, we applied PARL to these models. The results are displayed in Fig.~\ref{fig:ens_vgg16} and Fig.~\ref{fig:ens_lenet} respectively. For VGG16 we applied PARl to first 5 and 6 layers obtain a much higher robust accuracy compared to baseline with clean accuracy of 86.79\% and 82\% respectively. For LeNet-5, with only 2 convolution layers we apply PARL to first convolution and then both the convolution layers. We still observe a higher robust accuracy compared to baseline with clean accuracy of 71.67\% and 63.52\% respectively.

\begin{figure}[h]
\centering
    \includegraphics[width=\columnwidth]{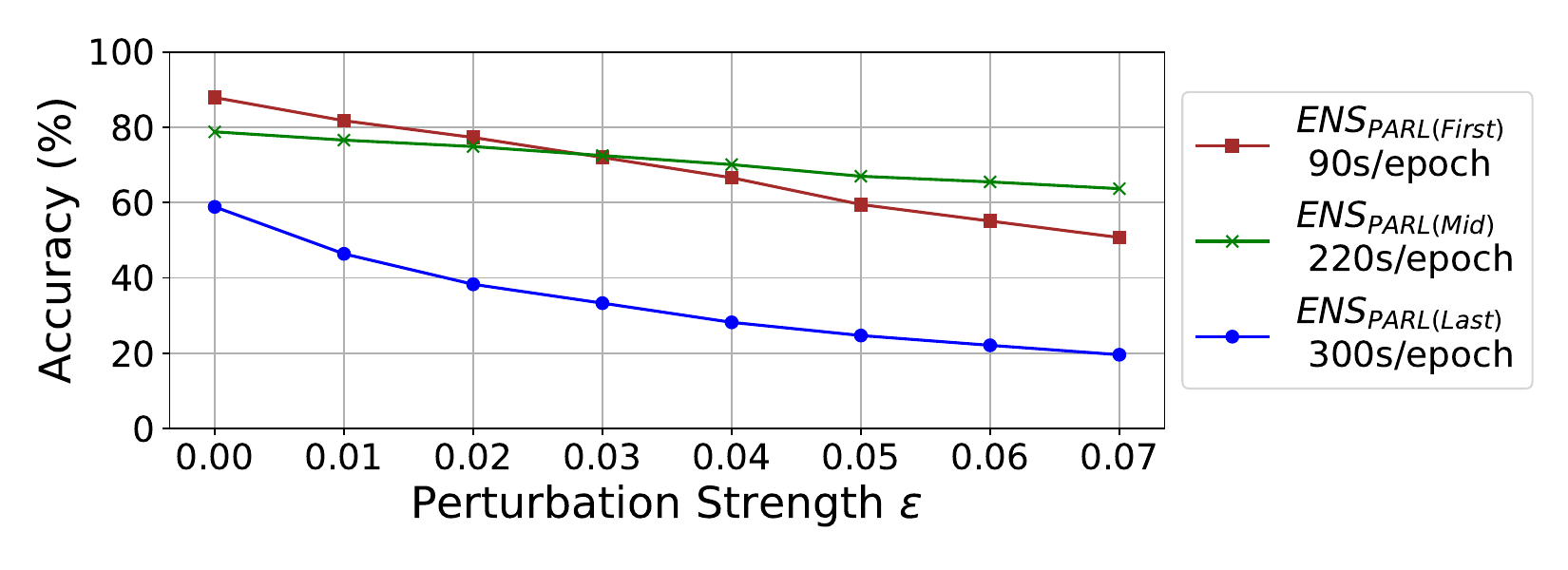}
    \caption{\footnotesize Resnet20 $ENS_{PARL/3/5}$ evaluation for CIFAR-10 with five layers selected from the start, middle, and end of the network \vspace{-0.4cm}}
    \label{fig:ens_mid_last}
\end{figure}

\section{Selection of initial, middle and last layers}
In Fig.~\ref{fig:ens_mid_last}, we present the clean and robust accuracy for $ENS_{PARL/3/5}$, with five convolution layers selected from the beginning, middle, and end of the network. We also include the per-epoch training time for each model. $ENS_{PARL(First)}$ achieves the highest clean accuracy, while $ENS_{PARL(Mid)}$ excels in robust accuracy, though with a slight decrease in clean accuracy. $ENS_{PARL(Last)}$ performs the worst in both clean and robust accuracy, likely because the final layers focus on converging the output, where introducing diversity is less effective. Overall, $ENS_{PARL(First)}$ offers the best trade-off between clean and robust accuracy, with the lowest training time.

\end{document}